\documentclass[nohyperref]{article}
\usepackage{microtype}
\usepackage{graphicx}
\usepackage{caption}
\usepackage{subcaption}
\usepackage{booktabs} 
\usepackage{amsmath}
\usepackage{amssymb}
\usepackage{amsmath,amssymb} 
\usepackage{paralist} 
\usepackage{fullpage}
\usepackage{multirow}
\usepackage{wrapfig}
\usepackage{inputenc}
\usepackage{color}
\usepackage{algorithm}
\usepackage{multirow}
\usepackage{amsthm}
\usepackage{thmtools}
\usepackage{xspace}
\usepackage{xcolor}
\usepackage{makecell}
\usepackage{epsfig}
\usepackage{microtype}
\definecolor{darkgreen}{rgb}{0.09, 0.75, 0.20}
\definecolor{amartya}{rgb}{0.09, 0.15, 0.90}
\definecolor{ameya}{rgb}{0.9, 0.45, 0.20}

\usepackage{enumitem}
\setlist{nosep}
\usepackage[pagebackref,breaklinks,colorlinks]{hyperref}
\hypersetup{
    colorlinks=false,
    linkcolor=blue,
    urlcolor=blue,
}

\usepackage[capitalize]{cleveref}
\crefname{section}{Sec.}{Secs.}
\Crefname{section}{Section}{Sections}
\Crefname{table}{Table}{Tables}
\crefname{table}{Tab.}{Tabs.}

\usepackage{comment}

\usepackage{graphicx}
\usepackage{amsmath,amssymb} 
\usepackage{color}
\usepackage{algorithm}
\usepackage{bbding}
\usepackage{pifont}
\usepackage{natbib}
\usepackage{capt-of}
\usepackage{amartya_ltx}
\newcommand{\ie}{\textit{i.e.}}

\usepackage{hyperref}
\usepackage{enumitem}

\PassOptionsToPackage{numbers,compress}{natbib}

\usepackage{amsmath}
\usepackage{amssymb}
\usepackage{mathtools}
\usepackage{amsthm}
\usepackage{authblk}

\usepackage[textsize=tiny]{todonotes}

\title{Towards Adversarial Evaluations for Inexact Machine Unlearning}

\date{}
\makeatletter
\renewcommand\AB@affilsepx{, \protect\Affilfont}
\makeatother

\author[1]{Shashwat Goel$^*$}
\author[2]{Ameya Prabhu$^*$} 
\author[3,4]{Amartya Sanyal} 
\author[5]{Ser-Nam Lim} 
\author[2]{Philip Torr}
\author[1]{Ponnurangam Kumaraguru}
\affil[1]{IIIT Hyderabad}
\affil[2]{University of Oxford}
\affil[3]{ETH Zurich}
\affil[4]{MPI-IS}
\affil[5]{Meta AI}

\begin{document}
\maketitle




\vspace{0.5cm}





\vspace{-1.35cm}
\begin{abstract}
    Machine Learning models face increased concerns regarding the storage of personal user data and adverse impacts of corrupted data like backdoors or systematic bias. Machine Unlearning can address these by allowing post-hoc deletion of affected training data from a learned model. Achieving this task exactly is computationally expensive; consequently, recent works have proposed \emph{inexact unlearning} algorithms to solve this approximately as well as evaluation methods to test the effectiveness of these algorithms.
    
    In this work, we first outline some necessary criteria for evaluation methods and show no existing evaluation satisfies them all. Then, we design a stronger black-box evaluation method called the Interclass Confusion (IC) test which adversarially manipulates data during training to detect the insufficiency of unlearning procedures. We also propose two analytically motivated baseline methods~(EU-$k$ and CF-$k$) which outperform several popular inexact unlearning methods. Overall, we demonstrate how adversarial evaluation strategies can help in analyzing various unlearning phenomena which can guide the development of stronger  unlearning algorithms.
\vspace{-0.3cm}
\end{abstract}


\section{Introduction}

Deep learning is becoming increasingly prevalent in everyday applications, with models being trained on large amounts of sensitive personal information including health and financial records, social network history, personal emails, and messages. This has led to growing privacy concerns, as codified in recent privacy legislation like GDPR~\citep{GDPR2018}, CCPA~\citep{CCPA2018}, and PIPEDA~\citep{PIPEDA2018}. The underlying motivation for privacy legislation is the concept of~\emph{data autonomy}, which states that every individual must retain complete control of their own data, including the right to withdraw their data from any system. However, deleting records corresponding to individuals is considerably harder for machine learning systems, especially those using deep networks, than with traditional databases.\vspace{0.1cm}

Recent studies, such as~\citep{feldman2020neural,zhang2017understanding}, have shown that deep neural networks have a tendency to \emph{memorize} data. This means that the network not only learns common patterns in the data, but also stores information about individual training data points. This is concerning from a privacy standpoint, as this information can be detected~\citep{shokri2017membership} or even extracted from the model~\citep{CarliniLEKS19}. The main goal of ``machine unlearning"~\citep{FTC21, Cao2015Unlearning, ginart2019deletion,
bourtoule2020machine, VILLARONGA2018Humans} is to design both algorithms to delete data stored in the network and evaluation methods to recover or detect the deleted data from the trained model.\vspace{0.1cm}

Preserving privacy and removing memorization are not the only motivations to study machine unlearning. Several studies have shown that small amounts of corrupted data can induce harmful properties into the trained model, which can greatly affect its behaviour on unseen
data \cite{nakkiran2020distributional}, a phenomenon we refer to as \textit{property generalization}. This can lead to problems in trustworthy machine learning, such as with noisy
data~\cite{frenay2014noise, northcutt2021confident, northcutt2021pervasive}, systematically biased
data~\cite{prabhu2020large}\footnote{to the extent that bias is a dataset problem~\cite{hooker2021moving}. \\ * equal contribution}, or adversarial data, such as poisoned samples~\cite{barreno2006security, Chen2017backdoor, Yang2017Poison}. For example,~\citet{sanyal2020benign,paleka2022law} show that a small amount of random noisy labels can significantly harm the adversarial robustness of a model. Further, ~\citet{konstantinov2022fairness} show that a small set of adversarially corrupted data can greatly increase unfairness. As these corrupted samples are discovered, unlearning them from the trained model can be used to remove the unwanted properties induced.\vspace{0.1cm}

Machine unlearning should remove both memorization and
generalized properties of deleted samples. This is captured in the concept of
\emph{model indistinguishability}, first defined
in~\citet{golatkar2019sunshine}. Let \(M\) be an ML model trained
on dataset \(S\) using learning algorithm \(T\) and \(S_f\subset S\)
be the set of points that need to be deleted from \(M\). 
 An unlearning process is considered successful if the 
distribution\footnote{due to
stochasticity in both the unlearning algorithm and \(T\)} of models produced by the unlearning process, is
indistinguishable from the distribution of models produced by
retraining the model using any training process \(T'\) on the
remaining data \(S\setminus S_f\). To see
why model indistinguishability implies unlearning, note that no training procedure
$T'$ which only uses $S \setminus S_f$ can produce a model that
carries information specific to $S_f$. Hence, it is sufficient to show the model distribution produced by the unlearning algorithm is indistinguishable from the model distribution produced by any one training algorithm \(T^\prime\) using $S \setminus S_f$. In this work, we study deleting a single query batch of samples. Extension to sequential deletion has to tackle challenges like correlated queries across
time~\citep{chourasia2022forget}.\vspace{0.1cm}

A naive method for unlearning data from a machine learning model is to
retrain the model on the retain data $S \setminus S_f$. This method
removes all information from the deletion set.  Hence, in theory, it
achieves ``exact unlearning", but is computationally and memory
intensive. Our work focuses on ``inexact unlearning", in which the
goal is to unlearn most information from the deleted data while
minimizing computational cost. While exact unlearning is often
infeasible, inexact unlearning presents a more tractable objective. In
the specific case of deep networks, due to the absence of theoretical
guarantees, empirical tests are commonly used for evaluating the
degree of unlearning. A strong empirical test should reliably distinguish models unlearning to varying degrees in terms of memorization and property generalization of the deletion set. The latter is challenging with existing evaluations which remove independently identically distributed (I.I.D) samples as the undesirable properties they induce may be apriori unknown.\vspace{0.1cm}

As comparing a distribution of ML models 
is intractable, most past evaluations compare the weights \cite{wu2020deltagrad,
izzo2021approximate} or outputs \cite{golatkar2019sunshine,
golatkar2020ntk, golatkar2021mixed, peste2021ssse} of an unlearnt model and one retrained using the original training procedure on $S \setminus S_f$. However, as we argue in~\Cref{thm:weight-not-suff}, even achieving
nearly identical weights is insufficient to guarantee similarity
in even well known properties like adversarial error and fairness, and thus model indistinguishability. This motivates
the need for adversarial evaluations of unlearning. We propose performing 
manipulation in the training data which introduces a known measurable property through
$S_f$ that is absent in $S \setminus S_f$. Thus, models that exhibit
this property cannot be indistinguishable from models obtained through
retraining on \(S\setminus S_f\). Ideally, the property unique to
$S_f$ should produce a large predictable change in model behaviour to
make its presence easy to measure. To this end, inspired by the
application of removing systematically biased data, we propose the
Interclass Confusion (IC) test. It induces the property of confusion
between two classes through label manipulations. IC test requires the
unlearning procedure to erase the induced confusion which we measure
as the number of samples of the two classes ``confused" as belonging
to the other class. As discussed in the following sections, we can use
this test to detect both memorisation and property generalisation in
an efficient way.\vspace{0.1cm}

We find that our proposed IC test is far stronger than existing
evaluations, allowing us to glean interesting insights into unlearning
algorithms.  Using the IC test, we can demonstrate the insufficiency
of a class of unlearning methods that simply modifies the final linear
layer~\cite{izzo2021approximate, baumhauer2020filtration} in deep
networks or methods that do not use the retained data $S \setminus S_f$ \cite{chundawat2022zero}. Our test detects the
presence of imperfectly unlearnt information about the deletion set $S_f$ in the early
layers of a deep network. Along with designing a stronger evaluation
method~(IC test), we also present two strong novel baselines --- EU-k,
which retrains the last \(k\) layers from scratch and CF-k, where a
model's last \(k\) layers are continually trained on the retain set
\(S\setminus S_f\). Finally, we also propose strategies to make the
original model $M$ more amenable to unlearning, thereby aiding faster unlearning. \vspace{0.1cm}

Overall, we emphasise empirical evaluations of inexact-unlearning
which measure how well an unlearning procedure forgets additional
properties induced by the deletion set $S_f$. Our work alleviates
certain shortcomings in existing evaluations as passing the IC test is
necessary for achieving model indistinguishability. Further, it's
adversarial nature makes it a much stronger test to pass than prior
evaluations as shown by our experiments. Our main contributions are: \vspace{0.1cm}

\begin{enumerate}
\itemsep0em 
    \item In line with the motivations of machine unlearning, we
    decompose the evaluation of unlearning into memorization and
    property generalisation. The former is computed on the forget set
    \(S_f\) whereas the latter is computed using unseen samples from the test set.
    \item We highlight some necessary principles for useful evaluations of unlearning not achieved in existing work. We alleviate this by introducing a
    new black-box evaluation called the Interclass Confusion~(IC)
    test. We empirically demonstrate that the IC test is far stronger
    than existing tests.
    \item Further, we use the IC test to show several surprising phenomena
    which may guide the design of future unlearning methods (i)
    Unlearning just the last layer only removes a small fraction of
    information about $S_f$ (ii) Unlearning methods may require the ability to learn \ie gain information (iii) Standard regularisation during
    training can make models more amenable to unlearning. Using
    these insights, we propose two strong baselines, EU-$k$ and CF-$k$, for comparing
    future unlearning methods. 
\end{enumerate}\vspace{0.1cm}

\textbf{Roadmap:} The rest of the paper is organized as follows: Section~\ref{sec:evaluations} describes our proposed evaluation methods in context of prior work, Section~\ref{sec:baselines} describes our unlearning baselines and their properties, Section~\ref{sec:experiments} presents our experimental results and in Section~\ref{sec:conclusion} we summarize our contributions while acknowledging the limitations of our work.

\vspace{-0.3cm}
\begin{table*}[t]
\vspace*{-0.2cm}
\caption{Comparison of evaluation methods (sampling strategy+metric) in inexact unlearning. Only our IC test satisfies all three desiderata.}\vspace{-0.1cm}
    \centering
    \resizebox{\linewidth}{!}{
    \begin{tabular}{l|c|ccc} \toprule
        \multirow{2}{*}{Deletion Set Sampling Strategy} & \multirow{2}{*}{Metric} & Necessary for  & Comparable Across  & Checks Property \\
        & & Indistinguishability  & Training Procedures & Generalization \\ \hline
        I.I.D \cite{golatkar2021mixed}, Class \cite{golatkar2019sunshine, golatkar2020ntk, golatkar2021mixed} & Relearn time & $\checkmark$ & $\times$ & $\times$\\
        I.I.D \cite{wu2020deltagrad, izzo2021approximate, thudi2021unrolling} & L2 Weights  & $\times$ & $\times$ & $\checkmark$\\
        I.I.D \cite{peste2021ssse} & L1-ConfusionMatrix & $\times$ & $\checkmark$ & $\checkmark$\\
        I.I.D \cite{golatkar2021mixed}, Class \cite{golatkar2020ntk, golatkar2021mixed} & L1-Softmax & $\times$ & $\checkmark$ & $\checkmark$ \\ \cline{1-2}
        \vtop{\hbox{\strut Class \cite{golatkar2021mixed,  golatkar2020ntk, baumhauer2020filtration},}\hbox{\strut I.I.D \cite{golatkar2021mixed, liu2021masking}}} & \multirow{2}{*}{MIA} & \multirow{2}{*}{$\checkmark$} & \multirow{2}{*}{$\checkmark$} & \multirow{2}{*}{$\times$}\\ \cline{1-2}
        \vtop{\hbox{\strut  I.I.D \cite{he2021deepobliviate, golatkar2021mixed, shibata2021CL+F},}\hbox{\strut Class \cite{golatkar2019sunshine, golatkar2020ntk, golatkar2021mixed}}} & \multirow{2}{*}{Error} & \multirow{2}{*}{$\times$} & \multirow{2}{*}{$\checkmark$} & \multirow{2}{*}{$\checkmark$} \\ \cline{1-2}
        Interclass Confusion (Ours) & MIA & $\checkmark$ & $\checkmark$ & $\times$ \\
        Interclass Confusion (Ours), I.I.D Confusion (Ours: Ablation) & Error & $\checkmark$ & $\checkmark$ & $\checkmark$ \\
        \bottomrule
    \end{tabular}}
    \vspace*{-0.25cm}
    \label{tab:test_properties}
\end{table*}

\section{Towards Adversarial Evaluations}
\label{sec:evaluations}
\vspace{-0.2cm}
We begin with an analysis of the shortcomings of prior evaluation strategies, and alleviate them by proposing the Interclass Confusion Test.

\vspace{-0.2cm}
\subsection{Recent Trends in Unlearning Evaluations}
\vspace{-0.15cm}

In this section, we look at existing approaches for evaluating unlearning procedures. First, these methods choose one of the two types of
deletion sets \(S_f\): $n$ I.I.D samples (I.I.D
removal)~\cite{golatkar2021mixed, wu2020deltagrad,
izzo2021approximate, peste2021ssse, he2021deepobliviate,
shibata2021CL+F} or $n$ samples belonging to a particular class (Class
Removal)~\cite{golatkar2019sunshine, baumhauer2020filtration}. Once
the unlearning procedure is applied on the above $S_f$, the
following are some popular metrics used to measure forgetting:\vspace{0.1cm}

\textit{Relearn Time}:~\citet{golatkar2019sunshine, golatkar2020ntk,
golatkar2021mixed, chundawat2022zero} measure the number of training
epochs until the loss of an unlearnt model drops below a pre-chosen
threshold when retrained on samples in $S_f$. A higher re-learn time
implies better forgetting.\vspace{0.1cm}

\textit{Weight Similarity}:\cite{wu2020deltagrad, izzo2021approximate}
measures the \(L_2\) distance between the weights of the unlearnt
model and another model retrained on \(S\setminus S_f\) using the
original training procedure. Naturally, a smaller distance is used to
imply better unlearning.\vspace{0.1cm}

\textit{Output similarity}: Similar to distance between weights,
distance in the softmax outputs on a pre-defined set of data points
are also used by different evaluation
methods.~\citet{golatkar2019sunshine, golatkar2020ntk,
golatkar2021mixed} measures the \(L_1\) distances between softmax
outputs of a unlearnt model and a retrained model on the pre-defined
set,~\cite{peste2021ssse} measures the \(L_1\) distance between the
confusion matrices of the respective
models,~\cite{he2021deepobliviate, golatkar2021mixed, shibata2021CL+F}
measure the gap in error on the distribution of affected samples. \vspace{0.1cm}

\textit{Membership Inference Attacks (MIA)}: Tests based on Membership
Inference Attacks~\citep{shokri2017membership,song2021systematic} are
designed to reliably distinguish data points in the training set from
similar unseen data. Hence, they can also be used to reliably
distinguish the deleted samples from similar unseen samples~(see
\citet{hu2021membership} for a survey). A detailed description of our
MIA attack compared to past unlearning literature is included in
Appendix Section~\ref{sec:MIA}.

\vspace{-0.2cm}
\subsection{Shortcomings of Existing Evaluations}
\vspace{-0.1cm}

We start by listing three desiderata absent in
most existing evaluation methods, as summarized in Table~\ref{tab:test_properties}. Theorem~\ref{thm:weight-not-suff} then motivates the need for adversarial evaluations.\vspace{0.1cm}

\begin{figure}[t]
\centering
\includegraphics[width=0.85\linewidth]{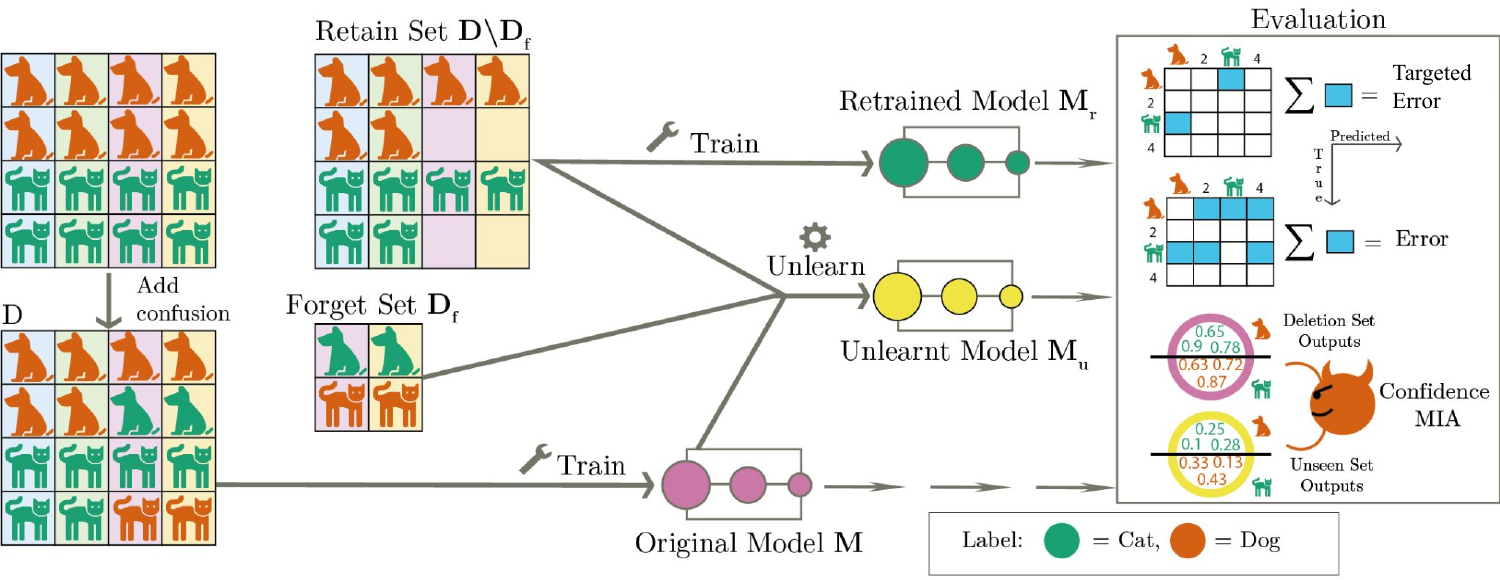}
\vspace*{-0.3cm}
\caption{IC Test Pipeline: We mislabel a subset of samples from two classes of the original dataset, forming $S_f$. Here, shape and colour represent the actual and labelled class respectively. Then, $M$ and $M_r$ are obtained by training from scratch on $S$ and $S \setminus S_f$ respectively. The unlearning procedure can leverage (some of) $M$, $S_f$ and $S \setminus S_f$ to produce the unlearnt model $M_u$.}
\vspace*{-0.45cm}
\label{fig:Pipeline}
\end{figure}

\textbf{Necessary for Indistinguishability}: Suppose there exists $T' \neq T$ such that retraining with $T'$ on $S \setminus S_f$ would produce a model highly similar to the unlearnt model $M_u$, then $M_u$ is a correct solution as it satisfies model indistinguishability. Showing no such $T'$ exists is difficult, and thus past evaluations simply compare $M_u$ with a model $M_r^T$ retrained using the original training procedure. However, this can exclude a large set of correct solutions which have no information from $S_f$ but behave differently from $M_r^{T}$. Consider a randomly initialized network. It clearly has no information from $S_f$, and indeed satisfies model indistinguishability if we consider $T'$ to be the random initialization process. However, it will be arbitrarily far from $M_r^T$ and will be unnecessarily penalized by evaluations based on Weights and Output similarity. Thus, passing past evaluations based on high similarity with a single model is not necessary for achieving unlearning.\vspace{0.1cm}

On the other hand, our proposed evaluation manipulates a subset of training data to introduce a measurable property through $S_f$ that is absent in $S \setminus S_f$. Any model that exhibits this property cannot be from a model distribution produced by (re)training without $S_f$ for all $T'$, and has not unlearnt. Thus, passing our evaluation is necessary to claim an unlearning procedure can handle arbitrary deletions.\vspace{0.1cm}

\textbf{Comparable Across Training Procedures}: Unlearning procedures
often significantly modify the training procedure or architecture
\cite{bourtoule2020machine, he2021deepobliviate, golatkar2019sunshine,
golatkar2020ntk, golatkar2021mixed, graves2020amnesiac}. Thus, a versatile unlearning evaluation should provide measurements of retained information that are comparable across chanes in architectures and training procedures. For example, measuring relearn time as an evaluation method
requires setting a threshold. However, different unlearning
procedures~\citep{golatkar2019sunshine, golatkar2020ntk,
golatkar2021mixed, chundawat2022zero} may differ in learning rate or
have inherently different behavior in how low the loss can get and how
fast it decreases. Similarly, $L_2$
distance between weights cannot be compared across architectures or hyperparameter choices like the amount of weight decay.\vspace{0.1cm}
 
\textbf{Checks Property Generalization}: Unlearning procedures must
ensure that properties which are only present in $S_f$ do not
influence performance on unseen samples. Some evaluations, such as
membership inference attacks (MIA) \cite{shokri2017membership,
chen2020machine} effectively only determine the removal of
memorization, and not the removal of generalized properties. In any evaluation with I.I.D removal, while it
is theoretically possible to check if generalized properties are removed, it is not clear what properties to look for.\vspace{0.1cm}

Two relevant properties that can be exacerbated by
corrupted data are adversarial
error~(\(R_{\mathrm{adv}}\))~\citep{madry2018towards} and
unfairness\footnote{We use accuracy discrepancy~\citep{buolamwini2018gender, sanyal2022unfair} for
mathematical simplicity. Can be shown for demographic parity and equalised
odds~\citep{hardt2016equality}.}~(\(\Gamma\)). An evaluation method that can be satisfied without removing these properties is clearly insufficient to guarantee unlearning. ~\Cref{thm:weight-not-suff} shows that metrics like $L_2$
distance in the parameter space and gap in test error on a random I.I.D
sample~(\(R\)) are poor indicators of whether two models have
similar \(R_{\mathrm{Adv}}\) and \(\Gamma\). \vspace{0.1cm}

\vspace{-0.15cm}
\begin{thm}\label{thm:weight-not-suff}
    There exists a distribution \(\cD\)  such that for any
    \(\epsilon, \alpha\geq 0\), there exist two \(\ell\)-layered fully
    connected linear NNs parameterised by \(\cW_1,\cW_2\) which are simultaneously:
    \begin{itemize}
        \item\textrm{\bfseries Close in Weights}:\; \(\norm{\cW_1-\cW_2}_F\leq \epsilon\)
        \item\textrm{\bfseries Close in Test Error}:\quad\(R\br{f_{\cW_1}}\leq
        R\br{f_{\cW_2}}+ \alpha\)
        \item\textrm{\bfseries Far in Robustness}:\;  \(R_{\mathrm{adv}}\br{f_{\cW_1}}\geq
        R_{\mathrm{adv}}\br{f_{\cW_2}}+ 1 - 2\alpha\)
        \item\textrm{\bfseries Far in Fairness}:\; \(\Gamma\br{f_{\cW_1}} =
        \Gamma\br{f_{\cW_2}} - 1\)
    \end{itemize}
    where \(R,R_{\mathrm{Adv},\Gamma}\) are as defined above and
    \(f_{\cW}\) is an \(\ell\)-layered fully connected linear neural
    network parameterised by \(W\). Proof is available in~\Cref{sec:proof-thm-1}.
\end{thm}
\vspace{-0.15cm}
The theorem shows that two models that are arbitrarily close in weights and test error can be arbitrarily far in adversarial robustness and fairness. In particular, we show an example where models get farther in adversarial robustness as they get closer in test error. Thus, unlearning evaluations must either measure indistinguishability in terms of more \emph{adversarial} quantities like robustness and fairness or use strategic non-I.I.D deletion sets. In our work, we explore the latter, i.e. an adversarial approach to designing deletion sets. In the following section, we introduce this evaluation procedure known as IC test. 

\vspace{-0.2cm}
\subsection{Proposal: Interclass Confusion Test}
\vspace{-0.1cm}

\label{sec:Tests}
 
In contrast to existing evaluations, we inject a strong differentiating influence specific to $S_f$ into the training dataset via label manipulations. Specifically, we present:\vspace{0.1cm}

\textbf{Interclass Confusion} (IC): As illustrated in Figure~\ref{fig:Pipeline}, the IC test using a deletion set of $n$ samples follows these steps:\vspace{0.1cm}

\begin{enumerate}
\itemsep0em
    \item Take $\frac{n}{2}$ samples each from two classes in the train data to form $S' \subset S$  (Targeted sampling).
    \item Swap labels \footnote{Note that evaluations based on label swapping have been used in traditional adversarial robustness literature \cite{nakkiran2019a, Fowl2021Poison}, but with quite different goals, setting and design.} between the two classes of samples in $S'$ (Adversarial manipulation) to get the confused set $S_f$. The dataset for training the original model $M$ is $(S \setminus S') \cup S_f$.
    \item Select the set $S_f$ as data to be deleted from the trained model $M$ (Strategic deletion set).
    \item Evaluate memorization and property generalization by measuring error on training and testing sets $S'$ and $S'_u$ corresponding to the two classes.
\end{enumerate}\vspace{0.1cm}

To isolate the effect of targeted sampling in the IC test \ie confusing two specific classes, we introduce: \vspace{0.1cm}

\textbf{Ablation: I.I.D Confusion}: We select $n$ samples uniformly at random from $S$ to form $S'$ and mislabel them to a uniform random different class, using these mislabelled samples as $S_f$.   Note that the removal is not I.I.D, we replace targeted label manipulation with I.I.D label noise.\vspace{0.1cm}

We compute the MIA and Error on affected classes like previous work, but also introduce the Targeted Error metric:\vspace{0.1cm}

\textbf{Error v/s Targeted Error}: Error computed for a given set $S$ is the fraction of samples in $S$ which were misclassified regardless of which class it was mistaken as. In Interclass Confusion, we are interested specifically in the fraction of samples confused between the two confused classes. In Class Removal, we are interested in the fraction of samples classified as the class to be removed. This is measured by Targeted Error, which is the fraction of samples in $S$ misclassified to the targeted class exhibiting the unwanted property is not removed. Samples misclassified into any other class are not counted as illustrated in Figure~\ref{fig:Pipeline} for IC test. As an illustrative example, for IC test on a 10 class dataset: the error of a random model would be 90\%, but the targeted error would be 10\%.
Error/Targeted Error when computed on the set $S'$ measures  memorization, and when computed on the unseen (test) set samples $S_t$ from the same distribution as set $S'$ measures property generalization.

\vspace{-0.3cm}
\section{Unlearning Baselines}
\label{sec:baselines}
\vspace{-0.15cm}
Having discussed properties of evaluation
methods, we now discuss unlearning procedures -- desirable properties and our two simple baselines that achieve
them.
\vspace{-0.3cm}
\subsection{Desiderata for Unlearning Methods}
\vspace{-0.1cm}

\textbf{Unlearning procedures need the ability to learn}: Consider a
linearly-separable binary classification task where we use the IC test
to introduce complete confusion between the two classes (50\% of samples of
each class mislabelled as the other). Powerful empirical risk
minimizers~(like neural networks trained with SGD) will achieve a
train accuracy on $S$ close to \(100\%\)~\citep{zhang2017understanding}.
However the test accuracy will be much lower, closer to \(50\%\), as
the training dataset is essentially fully randomly labelled.
However, upon deleting all the mislabelled samples, like in the IC
test, we are left with 50\% of the original dataset but with correct
labels. A model retrained from scratch on $S \setminus S_f$ can be expected to
achieve reasonably good accuracy, much larger than 50\%, which a good unlearning procedure is expected to match. So we can expect the unlearnt model to have learnt to perform the task, whereas the original model cannot. \vspace{0.1cm}

Intuitively, this implies that \emph{solely erasing information from
the model is not enough, and the ability to learn may be necessary for
ideal unlearning procedures}. Consequently, we expect methods which
do not use information about the retain set \cite{chundawat2022zero}
will have limitations when handling arbitrary deletions and will not perform well on the
IC test.\vspace{0.1cm}

\label{sec:scalabilitytolarge}
\textbf{Scalability to large deletion sets}: Popular unlearning
methods, both exact and inexact, either explicitly assume tiny
deletion sets \cite{thudi2021unrolling, bourtoule2020machine,
wu2020deltagrad} or scale poorly beyond them in practice
\cite{Schelter2020AmnesiaM, graves2020amnesiac, golatkar2019sunshine,
golatkar2020ntk}. In Appendix Section~\ref{sec:retrain_prob} we show
that the computational complexity of methods based on the paradigm of
isolating the influence of data to small parts of the training
procedure \cite{bourtoule2020machine, he2021deepobliviate,
ijcai2022arcane, graves2020amnesiac} scales exponentially with the
size of the deletion set. Arguably, methods which require resources
similar to retraining from scratch for large deletion sets
have limited practical value, especially in applications which require
large deletion sets~(see~\Cref{sec:retrain_prob} for a discussion). \vspace{0.1cm}

\textbf{Targeting Areas to Unlearn}: 
A way to significantly improve efficiency of unlearning procedures is
to focus optimization power towards areas of a model where the
deletion set is stored. We look from a layerwise perspective- the
early layers of a deep network capture generic low-level
representation \cite{yosinski2015understanding, kataoka2020pre},
while the later layers focus on dataset-specific information.
Interestingly, the earlier layers are also the most computationally
intensive~\citep{brock2017freezeout}. Hence, focusing unlearning on
the last $k$ layers may allow computationally efficient erasure of information from $S_f$. Such unlearning methods also help us analyze how early in a deep network can an evaluation method detect the presence of information specific to $S_f$.\vspace{0.1cm}

Overall, unlearning methods should: (i) have
capacity to learn information in addition to unlearning and (ii) scale
to large deletion sets and further, for our analysis, we wish to have
unlearning methods that (iii) target specific parts of the model, e.g. the last $k$
layers for unlearning.

\vspace{-0.25cm}
\subsection{Proposal for Unlearning baselines: CF-$k$ and EU-$k$}
\vspace{-0.1cm}

We propose two methods which we believe will be useful `baselines' for future work to compare against. (i) They achieve a tradeoff between forgetting and efficiency which can be controlled using parameter $k$, allowing comparisons with unlearning procedures of differing degrees of efficacy. (ii) They are simple and require minimal assumptions: they scale to large deletion sets, are applicable for all DNN training procedures and require only access to $S \setminus S_f$. \vspace{0.1cm}

\textbf{Exact-unlearning the last $k$ layers} (EU-$k$): We retrain the last $k$ layers of $M$ from scratch using the same training procedure $T$ on retain set $S \setminus S_f$ while freezing prior layers.  \vspace{0.1cm}

\textbf{Catastrophically forgetting the last $k$ layers} (CF-$k$): Neural Networks suffer from catastrophic-forgetting~\cite{FRENCHCatForget} - when a model is continually updated without some previously learnt samples, the model loses knowledge about them. We finetune the last $k$ layers of $M$ on the retain set $S \setminus S_f$ using the same training procedure $T$ while freezing prior layers, hoping to catastrophically forget $S_f$. As we avoid re-initializing the last $k$ layers unlike EU-$k$, we need far fewer epochs, making CF-$k$ more efficient than EU-$k$.\vspace{-0.4cm}

\section{Experiments}
\label{sec:experiments}
\vspace{-0.15cm}

We show empirical support for three claims of our work. (i) We show that our EU-$k$ and CF-$k$ unlearn better than four popular methods and are strong baselines. (ii) Using EU-$k$ and CF-$k$ for analysis, we show our primary contribution, the IC test, is more reliable than previous evaluations in detecting unwanted memorization and property generalization. (iii) We show standard regularization techniques can make original models $M$ more amenable to unlearning. Our training procedure is described in Appendix Section~\ref{sec:Implementation}. All code has been made publically available at \url{https://github.com/shash42/Evaluating-Inexact-Unlearning}.

\vspace{-0.2cm}
\subsection{Evaluating Baselines: EU-$k$ and CF-$k$}
\vspace{-0.1cm}

\textbf{Setup.} Due to the lack of established evaluation methods and comparisons to other methods in past work, the 'state of the art' in unlearning is not clear. We compare our proposed baselines against four popular past unlearning methods: Fisher \cite{golatkar2019sunshine}, NTK-Fisher \cite{golatkar2020ntk}, Amnesiac Unlearning \cite{graves2020amnesiac}, and LCODEC \cite{Mehta_2022_CVPR} which have published their codebase for accurate reproduction. We could not run Fisher, NTK-Fisher for our larger datasets like CIFAR10 due to large memory requirements and thus compare all models on their setting: We use Small-CIFAR-5 \cite{golatkar2019sunshine} (a 5 class subset of CIFAR10), and all samples of a given class as the $S_f$ (Class Removal). We follow their training procedure to get their original and retrain models, as we obtained near-random performance when we applied their unlearning method on our standard training procedure perhaps due to violation of some of their training assumptions. For Amnesiac, LCODEC and our unlearning procedures we report results on unlearning from an original model $M$ produced by our default procedure ($T$) with a standard ResNet-20 architecture. We obtain the same observations on using their respective training procedures which produce original models with lower accuracy. For forgetting we report memorization and property generalization by computing targeted error on the deletion set ($S_f$) and test set of the deleted class ($S_t$) respectively. We measure accuracy with test set error and efficiency with unlearning time.\vspace{0.1cm}
\begin{table}[t]
\vspace{-0.2cm}
\caption{Comparison between unlearning procedures on Class removal test on Small-CIFAR-5. Forgetting measured by targeted error: Memorization (Mem) and Property Generalization (PropGen). Performance and efficiency measured by test error and unlearning time. ($\downarrow$) indicates lower is better.}\vspace{0.1cm}
\centering
\resizebox{0.75\columnwidth}{!}{
    \begin{tabular}{lc|ccccc} \toprule\small
    Model & & \multicolumn{2}{c}{Targeted Error ($\downarrow$)} & Test Err ($\downarrow$) & Time(s) ($\downarrow$) \\ 
    & & Mem & PropGen & & \\ \hline
    \multicolumn{6}{c}{$T$ from \cite{golatkar2019sunshine,golatkar2020ntk}} \\ \hline
    Original & & 92.3 & 97.6 & 26.7 & 0.00 \\ \hline
 Fisher \citet{golatkar2019sunshine} &  & 94.6 & 98.0 & 33.2 & 141.95 \\
 NTK-Fisher \citet{golatkar2020ntk} &  & 27.0 & 39.6 & 31.0 & 141.90 \\ \hline
 Retrain &  & 0.0& 0.0 & 41.4 & 9.81 \\ \hline 
\multicolumn{6}{c}{$T$ (Ours)} \\ \hline
  Original &  & 98.0& 97.3 & 16.3 & 0.00 \\ \hline
  Amnesiac \cite{graves2020amnesiac} & & 22.3 & 21.6 & 74.3 & 1.72 \\ 
  LCODEC \cite{Mehta_2022_CVPR} & & 20.7 & 20.2 & 80.3 & 226.9 \\ \hline  
 \multirow{2}{*}{1-layer (Ours)} & CF & 18.3&12.3 & 30.9 & 4.43 \\
  & EU & 9.6 & 4.3 & 31.9 & 9.38 \\ \cline{1-2}
\multirow{2}{*}{10-layers (Ours)} & CF & 15.6 & 9.3 & 29.4 & 5.22\\
  & EU & 2.0 & 0.0 & 32.6 & 10.78\\ \hline
 Retrain &  & 0.0& 0.0 & 32.5 & 12.33 \\ \bottomrule
    \end{tabular}}\vspace{-0.3cm}
    \label{tab:Golatkar_CR}
\end{table}

\textbf{Results.} We present all our results in Table \ref{tab:Golatkar_CR}.\vspace{0.1cm}

\textit{Accuracy:} The test error of the retrained model is $15\%$ higher than the original (both $T$) because a portion of the test samples belong to the deleted class. Here, lower test errors are attributable to not forgetting the deleted class. Comparing test error for the Original models, we observe our procedure $T$ has a large decrease ($10\%$) in test error compared to \citet{golatkar2020ntk}. This ensures we study unlearning on better, more realistic models. We find that Amnesiac and LCODEC produce unlearnt models with almost random performance. Amnesiac relies on deletion set samples belonging to only a few batches. However, this assumption does not scale to large deletion sets and we find all batches are affected in our experiment, as expected from the mathematical analysis we present in Appendix Section~\ref{sec:against_isolation}. LCODEC removes samples sequentially, and the error of the model increases fast as more samples are deleted. \vspace{0.1cm}

\begin{figure}
\centering \vspace{-0.15cm}
    \includegraphics[width=0.85\linewidth]{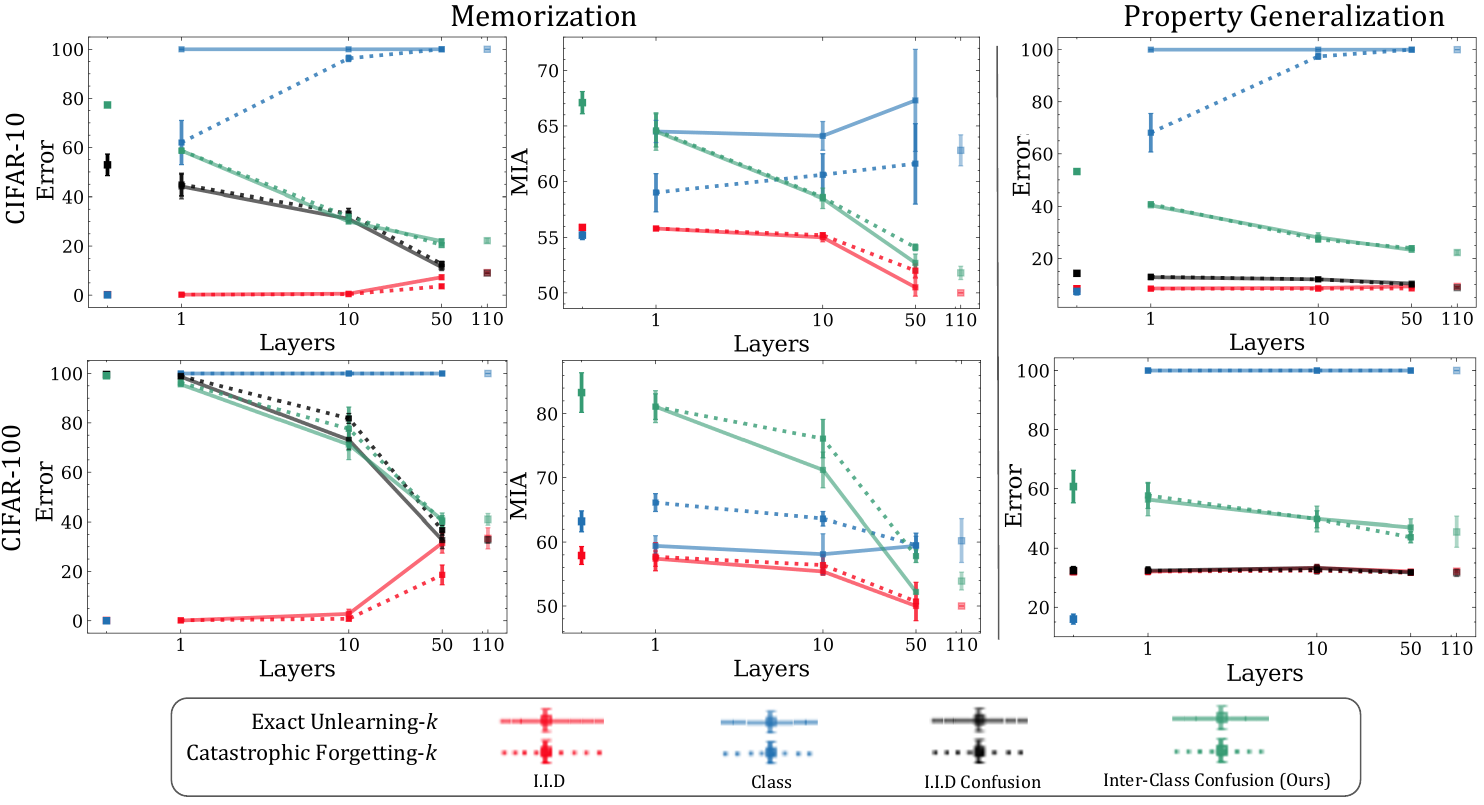}
    \vspace*{-0.25cm}
    \caption{Error, MIA for various deletion strategies (Y) reported across the number of layers (X) affected by the unlearning procedure. The left-most points at 0 layers represent the original model $M$, whereas the right-most points at 110 layers represent the retrained model $M_r^T$. Only Interclass Confusion reliably distinguishes different degrees of unlearning (no. of layers unlearnt) across all graphs.}
    \label{fig:evaluations}
    \vspace{-0.3cm}
\end{figure}
\textit{Forgetting:} We measure the degree of unlearning of a given method by comparing the reduction in targeted error of the method with the corresponding original and retrained models providing the starting and ideal scores respectively. We observe that simply unlearning the last layer with our baselines (EU-1 \& CF-1) have far better reductions in targeted error compared to previous methods in both memorization and property generalization. Surprisingly, \citet{golatkar2019sunshine} fails to achieve any significant forgetting. \vspace{0.1cm}

\textit{Efficiency:} We observe that 3 out of 4 past procedures take far more time for unlearning compared to our baselines and even retraining as approximating the Fisher Information Matrix is expensive. In real-world scenarios, such speedups are highly important to enable practical applications of unlearning. Amnesiac is fast but produces a random model.  \vspace{0.1cm}

\textbf{Conclusion.} Our methods $EU$-k and $CF$-k outperform popular unlearning methods by significant margins in all three dimensions: forgetting, accuracy and efficiency indicating they are reasonable baselines for analysis.

\vspace{-0.2cm}
\subsection{Comparing Tests for Evaluating Forgetting}
\vspace{-0.1cm}

\textbf{Setup.} We use CIFAR10 and CIFAR100 datasets with a 40K-10K-10K train-val-test split. Note that we use the same deletion set size $n$ for a fair comparison across all tests, with the sample set removed for every test, with details in Appendix Section \ref{sec:TestChoices}. Experiments in this section use $n$ corresponding to the number of training samples in one class: 4000 for CIFAR10 and 400 for CIFAR100 \cite{Krizhevsky09CIFAR}. We further report results across different deletion set sizes $n$ in the Appendix Section \ref{sec:untargetedvarying}, \ref{sec:targetedvarying} and find them to be consistent. All results are averaged over three runs with different seeds for robustness.  \vspace{0.1cm}

\textbf{Results.} In Figure~\ref{fig:evaluations} we compare different unlearning evaluation methods on their ability to demonstrate the degree of forgetting of models produced by our baselines EU-$k$ and CF-$k$. Every line is formed by varying the number of layers unlearnt $k$ and hence the degree of forgetting, with 0 and 110 (leftmost and the rightmost points) indicating the original and retrain models respectively. A strong test is indicated by: (i) the score of intermediate models ($0 < k < 110$) is different from that of the retrain model as some information is still retained after unlearning $k$ layers.
(ii) There is a clear gradual improvement in the forgetting metric as $k$ increases.
We present results consistent across graphs below:\vspace{0.1cm}

\textit{Memorization:} Class removal test (blue) is not able to detect memorized information even in simply exact-unlearning the last layer on any metric or dataset (solid-blue line reaches retrain scores immediately). I.I.D removal barely distinguishes 1-layer and 10-layer unlearning. We get random ($<$50\%) MIA scores for I.I.D confusion and hence exclude it. However, I.I.D confusion performs as well as the IC test on the error metric. IC test is the most useful across metrics and datasets, clearly distinguishing models with information removed from more layers. \vspace{0.1cm}

\textit{Property Generalization:} Only IC test is capable of detecting property generalization of confusion even after exact unlearning just the last layer. Even I.I.D confusion, which represents adding noisy labels with no systematic bias, is clearly insufficient to induce detectable generalized properties. Thus, both components of the IC test, class-targeted removal and confusion, are needed together to show clear trends in property generalization evaluations. \vspace{0.1cm}

\textit{Hyperparameters of IC Test:} The two hyperparameters in executing the IC test are choosing the two classes to confuse and the number of confused and deleted samples $n$. We find that while trends are similar across class pairs, unlearning is the hardest when we choose classes that are highly similar. We thus report results for (Cat, Dog) in CIFAR 10 and (Maple Tree, Oak Tree) in CIFAR 100 here. Further details can be found in Appendix Section~\ref{sec:varclasses}. Regarding the size of the deletion set, we demonstrate that the IC test can reliably detect imperfect memorization and property generalization with just 1\% and 5\% of the dataset being corrupted respectively. We report results for 5\% here, and show IC test is the most useful among all tests across deletion set sizes in Appendix Section~\ref{sec:targetedvarying}, \ref{sec:untargetedvarying}.\vspace{0.1cm}

\textit{EU-$k$ v/s CF-$k$:} For any given $k$, catastrophic-forgetting (CF) removes most of the information in those layers while being twice as fast as exact-unlearning (EU) indicated by the dotted lines closely following their solid counterparts across tests. The only cases where they differ is when the test is simply unable to detect information retained in exact unlearning (e.g. Class Removal). As shown by the IC test, EU-$1$ leaks a lot of information gained from $S_f$, showing that prior exact unlearning methods that only modify the final layer of deep networks \cite{baumhauer2020filtration, izzo2021approximate} continue to store information from $D_f$ and cannot handle arbitrary deletions \footnote{Only unlearning the final layer may succeed if earlier layers are trained privately \cite{guo2020certified, wu2020deltagrad}.}. Note that EU-$k$ and CF-$k$ continue to maintain the same accuracy across $k$ as shown in Appendix Table~\ref{tab:utilities_across_tests}. While we choose a 110 layer ResNet and $k=\{1, 10, 50\}$ as a concise representative sample here, all observations hold across more values of $k$ and network depths if compared using the fraction of layers unlearnt as shown in Appendix Section~\ref{sec:varlayers}. \vspace{0.1cm}

\begin{figure}[t]
    \centering
    \vspace{-0.25cm}
    \includegraphics[width=0.85\columnwidth]{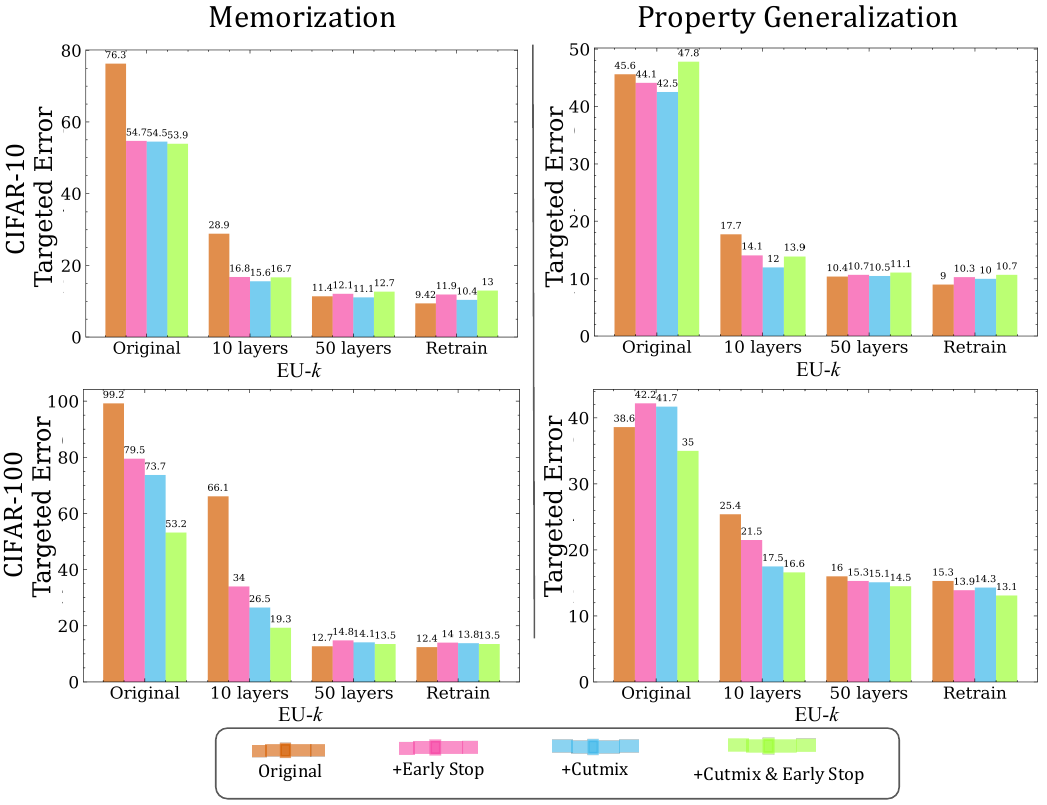}
    \vspace{-0.3cm}
    \caption{Interclass Confusion Targeted Error (Y) on unlearning from original models with different regularization (bar colors) reported for the original model $M$, EU-$10$, EU-$50$, and retrained model $M_r^t$. The same unlearning procedure can remove more confusion when starting from better regularized original models.}
    \label{fig:amenability}
    \vspace{-0.4cm}
\end{figure}
\textbf{Conclusion.} IC test is the only test that shows a clear difference between models with different number of layers unlearnt for both memorization and property generalizaton, on all metrics and datasets. IC test also shows past unlearning methods that propose to modify only the final layer of deep networks continue to retain most information about the deletion set. Varying $k$ in EU-$k$ and CF-$k$ can be used to control the forgetting-efficiency tradeoff at the same overall accuracy. Catastrophic forgetting achieved similar degree of forgetting as exact unlearning, while being twice as efficient.

\vspace{-0.2cm}
\subsection{Making Models Amenable to Unlearning}

\textbf{Aim.} Different original models $M$ can have varying propensities to memorize $D_f$. We aim to leverage this to provide training strategies that obtain original models $M$ with better unlearning properties, particularly computational efficiency. This is in line with recent work \cite{thudi2021necessity} which theoretically motivates this for $|S_f|=1$, but we empirically show it holds even for large deletion sets. \vspace{0.1cm}

\textbf{Strategies.} Early stopping has been a universal strategy to prevent overfitting (\ie memorization) in machine learning. We also use Cutmix~\cite{yun2019cutmix}, with the intuition that the model never sees a training sample in isolation while training, inspired from \cite{huang2020instahide}. Apart from using these regularization strategies during training, we use the same setup as before.\vspace{0.1cm}

\textbf{Results.} We present results in Figure \ref{fig:amenability}. Comparing original models (leftmost group of bars on the graphs), we observe that both techniques obtain large reductions in memorization of $S_f$ but similar property generalization. We observe only a marginal dropoff in unlearning (especially property generalization) from Cutmix+Early Stopping 10 layers to Original 50 layers. Cutmix+Early Stop 10 layers gives a huge improvement in unlearning performance compared to the Original 10 layers unlearnt models, especially on the harder CIFAR-100 dataset. In property generalization, this occurs despite original models having similar amounts of confusion indicating better regularized models make it easier for inexact unlearning methods to remove information.\vspace{0.1cm}

\textbf{Conclusion.} The presented results validate the idea that some original models make it easier to remove information using the same unlearning procedure. We demonstrated how this can be leveraged to achieve forgetting using cheaper unlearning procedures. Comparisons across unlearning procedures should ideally use the same original model for fairness, at least when there are no training assumptions.

\vspace{-0.3cm}
\section{Conclusion and Limitations}
\label{sec:conclusion}
\vspace{-0.1cm}

While prior unlearning methods claim to handle arbitrary deletion sets, we prove that passing prior evaluations based on weight and output similarity fail to guarantee unlearning of non-IID deletion sets. This motivates the need for adversarial evaluations like our proposed Interclass Confusion test. In contrast to prior evaluations, the IC test is necessary to pass to achieve model indistinguishability and is not sensitive to different training procedures. Even with a small fraction of data being manipulated, the IC test can reliably capture how well unlearning procedures remove memorization of deletion set samples and properties generalized from them -- both are important for different applications. We propose EU-$k$ and CF-$k$ as strong unlearning baselines that scale to large deletion sets, enable analysis of how early in the network information is retained and allow trading forgetting for efficiency at constant accuracy. We use our evaluation and methods to glean a variety of insights. (i) Unlearning methods that only modify the final layer in a deep network are not sufficient. (ii) We explore the interplay between learning and unlearning -- theoretically, we conjecture that an unlearning procedure aiming to handle arbitrary deletions requires the ability to learn. Empirically, we show that better regularized models are more amenable to unlearning.\vspace{0.1cm}

We hope that our analysis and proposed IC test along with EU-k and CF-k baselines will enable building stronger adversarial tests and better unlearning procedures. There is a need to bridge the current limitations of our work: We do not expect EU-$k$ and CF-$k$ to be gold-standard unlearning procedures, they are meant only as simple analytical tools that assist future research. As for the IC test, defining a passing score for real-world datasets that is necessary and sufficient is an open problem. While any procedure claiming to handle arbitrary deletions must pass the IC test, it alone cannot guarantee perfect unlearning. Finding a test that if passed is sufficient to prove unlearning of arbitrary deletions is an interesting direction. \vspace{0.1cm}

Finally, we hope our work spurs the use of unlearning in applications like removing systematic bias, noise, and tackling adversarial manipulations. 

\vspace{0.1cm}

\textbf{Acknowledgements:} We thank Shyamgopal Karthik, Saujas Vaduguru, Arjun T.H., Naren Akash RJ, Shradha Sehgal, Nikhil Chandak, Shashwat Singh, Shashwat Chandra among others for helpful feedback. Ameya Prabhu was funded by Facebook Grant Number DFR05540. This work is supported by the UKRI grant: Turing AI Fellowship EP/W002981/1 and EPSRC/MURI grant: EP/N019474/1.

\bibliographystyle{icml2023}
\bibliography{paper}
\hfill \break

\appendix
\onecolumn
In the appendix, we provide additional information and experiments to supplement the observations in the main paper. The appendix is organized sectionwise as follows: 
\begin{enumerate}[label=(\Alph*)]
    \item We detail all notations and abbreviations used in the main paper.
    \item We prove Theorem~\ref{thm:weight-not-suff}, analyze data isolation strategies, and discuss membership inference attacks.
    \item We provide details of implementation, test and metric choices, unlearning method comparisons and utility calculations.
    \item We show empirical results on (i) varying the number of unlearning epochs, (ii) layers in the architecture and (iii) samples to be deleted. We also vary the choice of confused classes and ablate the effect of warm restarts in our training procedure. These results demonstrate the robustness of our observations to changes in optimization details, model, selected classes, and training procedure.
\end{enumerate}

\section{References: Notations and Abbreviations} 
In Tables~\ref{tab:notations} and \ref{tab:terms} we list the notations and terms used in this work respectively for reference when reading the paper. 
\begin{table}[t]
\centering
\vspace*{-0.35cm}
    \caption{Reference for notations used in this work}
\resizebox{0.9\textwidth}{!}{ 
    \centering
    
    \begin{tabular}{ll} \toprule
       Abbr.  & Definition  \\ \hline
$T$ & The default training procedure. Implementation listed in Appendix Section~\ref{sec:Implementation}.\\ 
$S$ & The entire training set.\\ 
$S_f$ & The samples to be removed, called the deletion set.\\ 
$S_t$ & Unseen (test set) samples from the same distribution (here, affected classes) as the deletion set.\\ 
$M$ & The original model, obtained from  the training procedure $T$ using the entire train set $S$.\\ 
$M_u$ & The unlearnt model, obtained from applying some unlearning procedure on $M$.\\ 
$M_r^T$ & The retrained model, obtained from the same training procedure $T$ using the retained data $S \setminus S_f$. \\ 
$\phi_u$ & The distribution of models obtained from applying an unlearning procedure on the original model $M$. \\ 
$\phi_r$ & \vtop{\hbox{\strut The distribution of models obtained by retraining from scratch}\hbox{\strut using a training procedure $T'$ on the retain set $S \setminus S_f$.}} \\ 
$P$ & Number of parts a data \& influence isolation strategy divides the training data.\\ 
$S'$ & The subset of the original dataset $S$ selected for manipulations in unlearning testing. \\ 
$S'_u$ & The subset of the test set (unseen samples) from the same distribution (in our case, affected classes) as $S'$. \\
$A, B$ & The confused classes in the IC test.\\ 
$n$ & The size of the deletion set to be chosen. \\ \bottomrule
    \end{tabular}}
    
    \label{tab:notations}
\end{table}

\begin{table}[t]
\centering

\caption{Reference for abbreviations used in this work}
\resizebox{0.6\columnwidth}{!}{ 
    \centering
    
    \begin{tabular}{ll} \toprule
       Abbr.  & Definition \\ \hline
IC & Interclass Confusion test \\  
MIA &  Membership Inference Attack(s) \\  
CF-$k$ & Catastrophic-Forgetting $k$ layers \\ 
EU-$k$ & Exact-Unlearning $k$ layers \\ 
DNN & Deep Neural Network \\ 
I.I.D & Independent and identically distributed \\ 
NTK-Fisher & Neural Tangent Kernel + Fisher method of \cite{golatkar2020ntk} \\ 
stdev & Standard Deviation \\ 
\bottomrule
    \end{tabular}}
    
    \label{tab:terms}
    \vspace*{-0.25cm}
\end{table}


\section{Additional Analysis}

\subsection{Proof of Theorem 1}
\label{sec:proof-thm-1}

\begin{proof}
    
    We prove this by constructing two \(\ell\)-layered fully connected
    linear NNs, parameterised by \(\cW_1,\cW_2\) and a distribution
    \(\bP\) such that, under \(\bP\) they are close in weights and
    test error but far in robustness and fairness.

    \noindent Let \(\cW_1=\{A^1,\ldots,A^\ell\}\) and \(\cW_2=\{B^1,\ldots,B^\ell\}\) be the list of weight matrices of the two networks with each matrix having a dimension of \(m\times m\). Consider all but the first layer of the two networks be identical. Specifically, 
    $$
    A^2 = A^3 \dots = A^l = \begin{bmatrix}
        \sigma & & & & \\
        & \sigma & & & \\
        & & \ddots & & \\
        & & & & \sigma & \\
        & & & & & \sigma_{1}
        \end{bmatrix}
    $$ with the remaining entries being \(0\) where we will define \(\sigma,\sigma_1\) later.
    Construct the first layer of the two networks as follows where \(\epsilon>0\).
    $$
    A^1 = \begin{bmatrix} 
        1 & & & & \\
        & 1 & & & \\
        & & \ddots & & \\
        & & & & 1 & \\
        & & & & & 0
        \end{bmatrix}
    \qquad
    B^1 = \begin{bmatrix} 
        1 & & & & \\
        & 1 & & & \\
        & & \ddots & & \\
        & & & & 1 & \\
        & & & & & \epsilon
        \end{bmatrix}
    $$
    \paragraph{Closeness in $L_2$ weights} By construction, the two neural networks are close in weights: $\norm{\cW_1-\cW_2}_1= \epsilon$ where \( \norm{\cW_1-\cW_2}_1=\sum_{i=1}^\ell \norm{A^i-B^i}\).

    \noindent To complete the remainder of the proof, note that the two neural networks are essentially equivalent to linear functions with the weight parameters \(A\) and \(B\) respectively where
    \[A = \begin{bmatrix} 
        \sigma^{l-1} & & & & \\
        & \sigma^{l-1} & & & \\
        & & \ddots & & \\
        & & & & \sigma^{l-1} & \\
        & & & & & 0
        \end{bmatrix}
    \qquad
    B = \begin{bmatrix} 
        \sigma^{l-1} & & & & \\
        & \sigma^{l-1} & & & \\
        & & \ddots & & \\
        & & & & \sigma^{l-1} & \\
        & & & & & \sigma_1^{l-1}\epsilon
        \end{bmatrix}\]
    
    \noindent Next, we construct a data distribution \(\bP\) that satisfies the criteria of our result. Our distribution \(\bP\) will be supported on four points \(X^1,X^2,X^3,X^4\in\reals^m\) where \[X^1=\br{\underbrace{1,\ldots,1}_{m-1}, 0}, X^2=\br{\underbrace{-1\ldots,-1}_{m-1}, 0}, X^3=\br{\underbrace{1,\ldots,1}_{m-1}, -1}, X^4=\br{\underbrace{-1,\ldots,-1}_{m-1}, 1}\] and \(\bP\) is defined as \[\bP[(X^1,+1)]+\bP[(X^2,-1)]=1-\alpha~\text{and}~\bP[(X^3,+1)]+\bP[(X^4,-1)]=\alpha.\]
    
    \paragraph{Test Error} It is easy to verify that if \(\sigma>0\),
    then \(R(f_{\cW_1}), R(f_{\cW_2})\leq \alpha \).

    \paragraph{Fairness} Now, let \(\bc{X^3\cup X^4}\) be the minority
    group and \(\bc{X^1\cup X^2}\) be the majority group. Note that
    \(f_{\cW_1}\br{X^1}=f_{\cW_1}\br{X^3}=1\) and
    \(f_{\cW_1}\br{X^2}=f_{\cW_1}\br{X^4}=-1\), thereby leading to
    \(\Gamma(f_{\cW_1})=0\). On the other hand, for any \(\epsilon,m\)
    if \(\sigma,\sigma_1\) are chosen such that 
    \begin{equation}
        \label{ineq:unfair}
        \sigma_1^{l-1}\epsilon> (m-1)\sigma^{l-1},
    \end{equation} we have that \(f_{\cW_2}\br{X^3}=-1\) and
    \(f_{\cW_2}\br{X^4}=1\). Hence, \(\Gamma\br{f_{\cW_2}}=1\). This
    completes the proof of \[\Gamma\br{f_{\cW_2}} -
    \Gamma\br{f_{\cW_1}}=1.\]

    \paragraph{Adversarial Robustness} Let \(\delta>0\) be the
    adversarial perturbation budget. Then, the adversarial error of a
    network parameterised with parameters \(\cW\)
    is\begin{align*}
        R_{\mathrm{Adv}}(f_{\cW})&=\bP_{X,y}\bs{\exists\vec{z}\in\reals^m~\text{s.t.}~\norm{\vec{z}}\leq\delta~\wedge~f_{\cW}\br{X+\vec{z}}\neq
    y}\\
    &\geq \alpha \bI\bc{\exists
    \vec{z}_1,\vec{z}_2\in\reals^m~\text{s.t.}~\norm{\vec{z}_1},\norm{\vec{z}_2}\leq\delta~\wedge~f_{\cW}\br{X^1+\vec{z}_1}\neq
    1~\wedge~f_{\cW}\br{X^2+\vec{z}_2}\neq -1}.  \end{align*}    Note
    that if the parameters \(\sigma,\sigma_1\) satisfy the following
    with respect to \(\epsilon,\delta\)
    \begin{equation}\label{eq:adv-robust-param}
    \sigma_1^{l-1}\epsilon\delta> (m-1)\sigma^{l-1} \end{equation}
    then \(f_{\cW_2}\br{X^1-\delta e_{m}}=-1\) and
    \(f_{\cW_2}\br{X^2+\delta e_{m}}=1\) where \(e_m\) is the \(m^{\it
    th}\) canonical basis vector. Thus,
    \(R_{\mathrm{Adv}}(f_{\cW_2})\geq 1-\alpha\). It is also easy to
    verify that for any \(\vec{z}:~\norm{\vec{z}}\leq \frac{1}{2}\),
    we have that for \(f_{\cW_1}\br{X^1+\vec{z}}=1\) and
    \(f_{\cW_1}\br{X^2+\vec{z}}=-1\).  Thus,
    \(R_{\mathrm{Adv}}(f_{\cW_1})\leq\alpha\). Subtracting the two
    adversarial errors we obtain,
    \(R_{\mathrm{Adv}}(f_{\cW_1})-R_{\mathrm{Adv}}(f_{\cW_2})\geq
    1-2\alpha\).

    \noindent Finally, combining~\Cref{eq:adv-robust-param,ineq:unfair}  and setting the parameters such that \(\frac{\sigma_1}{\sigma}\geq \br{\frac{m-1}{\epsilon\delta}}^{\frac{1}{l-1}}\) completes the proof.
\end{proof}

\subsection{Against Isolation Strategies}
\label{sec:against_isolation}
\begin{figure}[t]
\centering
\includegraphics[width=0.65\linewidth]{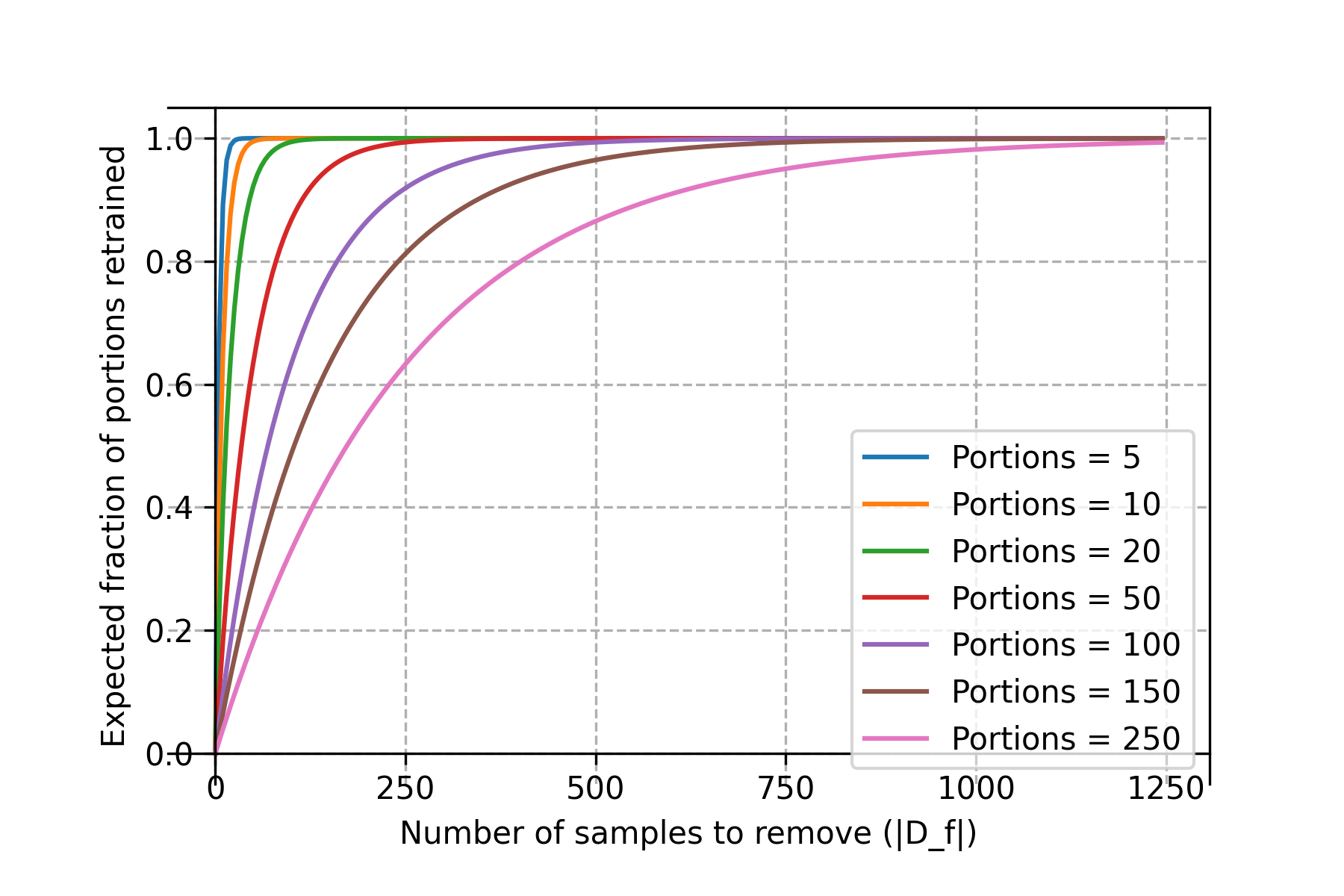}
\vspace*{-0.3cm}
\caption{Hyperbolic deterioration of efficiency in isolation-based unlearning when scaling to a large number of removed samples. In this work, we analyze $|S_f|$ from 100-4000 where $\mathbb{E}[Y] \sim 1$.}
\label{fig:isoeff}
\vspace{-0.35cm}
\end{figure}

Examples include removing noisy labels \cite{northcutt2021confident, northcutt2021pervasive}, deleting poisoned samples \cite{Wang2019Cleanse, jagielski2018manipulating, li2020backdoor}, deleting data that induces harmful biases \cite{prabhu2020large, fabbrizzi2021surveybias}, and organizations requiring deletion of user data older than some retention period. Even in the context of privacy, a single user might \emph{own} multiple samples in the dataset. In biometrics like face recognition \cite{turk1991eigenfaces}, one user may form an entire class \cite{baumhauer2020filtration}. Moreover, user deletion requests may occur in bursts after certain \emph{events of interest}, such as revelations of privacy leakages by an organization \cite{acquisti2006breach}. Lastly, batching online deletion requests requires less invocations of the unlearning procedure, boosting resource efficiency. 

A popular approach for unlearning is data-influence isolation, where each sample is made to contribute only to a small part of the training procedure or model. Unlearning such as retraining from scratch only the part affected by the deletion set erases the influence of the deletion set more efficiently. Isolation-based strategies change the training process by creating an ensemble~\cite{ijcai2022arcane, Schelter2020AmnesiaM, bourtoule2020machine, graves2020amnesiac, he2021deepobliviate}, each of whose models is trained on different subsets of the dataset. This ensures architecturally \cite{bourtoule2020machine, Aldaghri_2021, Schelter2020AmnesiaM, ijcai2022arcane} or temporally \cite{he2021deepobliviate, bourtoule2020machine} isolating the influence of any sample to a limited part of training, requiring retraining for only the affected parts. Isolation has been used across techniques like Linear Classification \cite{Aldaghri_2021}, Random Forest \cite{Schelter2021Hedgecut, brophy2021forests}, KNN \cite{COOMANS1982KNN}, SVM \cite{Cauwenberghs2003SVM, tsai2014linear} and DNN \cite{graves2020amnesiac, bourtoule2020machine, he2021deepobliviate} by utilizing or creating a sparse influence graph \cite{Schelter2020AmnesiaM}. Data-influence isolation often comes at the cost of utility as each portion becomes a weaker learner \cite{Banko2001Large}, especially in deep networks \cite{shorten2019survey}. To overcome the dropping utility, the training and unlearning time may need to be increased, reducing resource efficiency.

Figure~\ref{fig:isoeff} demonstrates that the computation costs of isolation-based strategies scale poorly as the deletion set size increases. Note that even on a practical deletion set size like 500, existing isolation based approaches (which create much less than 250 isolated portions) require almost full retraining costs on expectation. Let $P$ be the number of parts obtained with the isolation strategy. We assume the best-case scenario where each sample only influences one part. We make the simplifying assumption that the samples are uniformly distributed across parts, and the probability of a removed sample belonging to any particular portion remains constant ($\frac{1}{P}$). Let $Y$ be the number of affected parts. The probability part $i$ is affected by atleast one sample in $S_f$ is $1 - (1 - \frac{1}{P})^{|S_f|}$. Thus by the linearity of expectation: $\mathop{\mathbb{E}}[Y] = P\left(1 - \left(1 - \frac{1}{P}\right)^{|S_f|}\right)$.

\begin{figure}[t]
\centering
\includegraphics[width=0.65\linewidth]{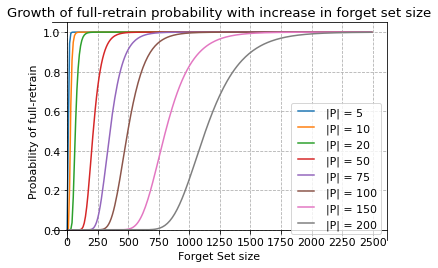}
\caption{Logistic growth of the probability of needing to retrain all portions with increasing deletion set size. We represent isolation strategies with different portion sizes $P$. }
\label{fig:retrain_prob}
\vspace{-0.5cm}
\end{figure}

\label{sec:retrain_prob}
We also show that the probability of full-retrain in data-influence isolation unlearning methods scales poorly with increasing deletion set size. Let $p(n)$ be the probability that all $P$ portions are affected on the removal of $n$ samples. Extending the analysis of~\citet{warnecke2021machine} from the specific case of SISA to data-influence isolation in general, we get:
\begin{equation*}
p(n) = 1 - \frac{\sum_{j=1}^{|P|} (-1)^{j+1} \binom{|P|}{j} (|P|-j)^n}{|P|^n}
\end{equation*} 
Figure~\ref{fig:retrain_prob} shows $p(|S_f|)$ grows logistically, implying there is a fast increase in the chance of needing a full-retrain as deletion sets get larger. This demonstrates how data-influence isolation provides little improvement in efficiency compared to the retrain-from-scratch baseline for practical scenarios.

Note that some prior work such as \cite{graves2020amnesiac} do not re-train the affected portions without the deleted data, instead removing them entirely. This replaces the hit on resource efficiency with decreased utility (such as accuracy) as more deletions lead to more affected portions being removed from the model. This explains why \cite{graves2020amnesiac} produces an almost random model in the Class Removal experiment shown in Table~\ref{tab:Golatkar_CR}.

\subsection{Membership Inference Attacks}
\label{sec:MIA}

\subsubsection{Background}
Membership Inference Attacks (MIA) \cite{shokri2017membership} can be used to determine whether a particular sample was part of the training data of a model. Many different black-box formulations of MIA have been used to measure the efficacy of unlearning. Most \cite{golatkar2020ntk, golatkar2021mixed, liu2021masking, graves2020amnesiac} learn a binary attack classifier: based on the model's output for the sample, was the sample in the seen training set (class 0) or the unseen test set (class 1)? The attack classifier is then applied on deletion set samples, with ideal unlearning entailing all samples are classified as unseen. However, such a test is extremely sensitive to the efficacy of the attack classifier which may be unreliable. Another approach has been to train the attack classifier to distinguish the outputs of a large number of original ($M$) and retrained ($M_r^T$) models and then classify the unlearnt model $M_u$ \cite{baumhauer2020filtration}. This formulation involves prohibitive computational expense and still can't check over all potential $T' \neq T$, indistinguishability with any of whom would guarantee unlearning.

\citet{song2021systematic} recently show that \emph{metric-MIA}, measuring simple metrics and deciding membership based on a threshold, can match the classification accuracy of trained attack models. In particular, their confidence-based MIA measures the model's output probability for the target class and selecting separate class-wise membership thresholds. It is shown to match the performance of even white-box MIA attack classifiers.

\subsubsection{Our Formulation}
\label{sec:OurMIA}

We adapt the confidence-based MIA \cite{song2021systematic} to propose an efficient black-box MIA formulation specifically tailored for measuring forgetting. We assume direct access to the actual model outputs instead of shadow models \cite{graves2020amnesiac, liu2021masking}, as shadow models only weaken the attack, making the unlearning test artificially easier to pass. We distinguish the model outputs on samples from the deletion set $S_f$ and unseen samples $S_t$ from the same underlying distribution rather than training an attack classifier using the entire train and test set. We believe this formulation is a more targeted measurement of forgetting as it directly discriminates between outputs on $S_f$ and $S_t$ in contrast to train and test set used in past literature~\cite{golatkar2020ntk, golatkar2021mixed, graves2020amnesiac, liu2021masking}. 

Our MIA takes in model $M$, forget set $S_f$ and unseen samples $S_u$ from the same classes found in $S_f$. The following procedure is repeated for each 'target class' $t$:

\begin{itemize}
    \item Dataset $S_{MIA}$ is created with the probability outputs for class $t$: $M(S_f)_t$ and $M(D_u)_t$ stored as class 0 and class 1 respectively.
    \item We then create a 50-50\footnote{Given that only 1 parameter (threshold) needs to be learnt, the shadow size is sufficient} shadow ($S_{MIA-S}$) - test ($S_{MIA-T}$) split of $S_{MIA}$.
    \item A threshold $p_t$ needs to be chosen such that probabilities $> p_t$ are classified as class 0, and probabilities $< p_t$ as class 1. The $p_t$ that maximizes the accuracy on $S_{MIA-S}$ is chosen.
    \item The accuracy obtained on $S_{MIA-T}$ using threshold $p_t$ is the MIA accuracy for target class $t$. A weighted average of this test accuracy across all target classes is taken as the final MIA accuracy.
\end{itemize}

Usually, the target class $t$ is the actual label of the sample. However, in the case of IC test, we use the mislabelled class as the target for both, $S_f$ and $S_u$ samples. Intuitively, the memorization of mislabels in the deletion set would make the wrong class probability output unnaturally higher than other unseen samples of the same class, making the MIA stronger. Such an enhancement is not possible in the case of I.I.D confusion as the mislabels are untargeted. 

In line with existing MIA literature, we want our attack classifier accuracy to be 50\% incase of no classifier advantage. Thus as the forget set and unseen set may have differing sizes in some experiments, we take a random subset of the larger one to make the attack dataset balanced. The numbers reported are averaged over 20 runs with randomness induced by the subset sampling step. Note that since the classifier learns to distinguish between the test and forget set distribution directly, it might be able to distinguish them spuriously, leading to slightly more than 50\% attack classifier accuracy even on perfect unlearning. Thus, the reference gold standard MIA performance can instead be that of any exactly unlearnt model upon undergoing the same evaluation.

\section{Additional Details}

We now provide some additional details for results shown in the main paper.

\subsection{Implementation Details}
\label{sec:Implementation}

\textbf{Training.} We use the ResNet architecture \cite{he2015deep} with 110 layers. Our standard training procedure $T$ is as follows: We train our models for 62 epochs (CIFAR10) or 126 epochs (CIFAR100), using a SGD optimizer with momentum 0.9 and weight decay 5e-5, an SGDR scheduler with $t_{mult}$ = 2, $t_0$ = 1, minlr = 5e-3, maxlr = 0.01 and a batch size of 64. For EU-$k$ and CF-$k$ baselines, we use this same training process, but on the final $k$ layers. In CF-$k$, the only difference is we finetune for only half the epochs.  

The setup used for all experiments is a PC with a Intel(R) Xeon(R) E5-2640 2.40 GHz CPU, 128GB RAM and 1 GeForce RTX 2080 GPU. 

We make the following deviations in our experiments:
\begin{itemize}
\item In Table~\ref{tab:Golatkar_CR} we make changes described in Section~\ref{sec:baselines}.
\item In Figure~\ref{fig:amenability} and Table~\ref{tab:regularization_utilities} we change the training procedure. When using cutmix regularization, we use $p=0.5$ and $\alpha=1.0$. For early stopping, we halve the number of epochs both while training the original/retrain models and also in the unlearning procedures.
\item In Table~\ref{tab:varyingCFepochs} we vary the number of finetuning epochs in CF-$k$.
\item In Figure~\ref{fig:varclasses} we vary the confused classes in the IC test from easy-hard on the axis of distinguishability.
\item In Figure~\ref{fig:varlayers} we further benchmark on ResNet-20, ResNet-56 and ResNet-110 to show our results are robust to the choice of network depths.
\item In Table~\ref{tab:norestarts} we ablate the effect of warm restarts in training the original/retrain model.
\end{itemize}

\subsection{Metrics}
\subsubsection{Inclusion in the Evaluation Comparisons Table}
Note that the list of metrics in Table 1 of the main paper does not include metrics like upper bound on information remaining in weights and activations \cite{golatkar2019sunshine, golatkar2020ntk, golatkar2021mixed} since its unclear whether such metrics can be computed on methods other than their own proposed unlearning procedure. We also exclude purely-qualitative tests such as model inversion attacks \cite{fredrikson2015model} which have been used in prior unlearning works \cite{graves2020amnesiac, baumhauer2020filtration}.

\subsubsection{Details of Metric Computation}
\label{sec:metric_comp}

\textbf{Targeted Error} We propose Targeted Error which measures the number of samples classified according to a property (information) unique to $S_f$. For the IC-test, it is the fraction of samples still confused between the two classes, i.e. $ \texttt{Targeted Error}(M, S, A, B) = \frac{C^{M,S}_{A,B}+C^{M,S}_{B,A}}{|S_A| + |S_B|} $. where $C^{M,S}$ is the confusion matrix when using model $M$ outputs on dataset $S$ and $A, B$ are the classes confused. \\
For confusion between $N>2$ classes, targeted error is the sum of the confusion matrix terms for all pair-wise misclassifications among the $N$ classes. Thus, targeted error converges to error when $N$ is the same as the total number of classes, as in I.I.D Confusion. For Class Removal test, targeted error calculates the number of samples labelled as the removed class.

Note that the influence of utility on targeted error is significantly lesser than the simple error metric on the affected classes as illustrated in Figure~\ref{fig:Pipeline}. Regarding the passing score for IC test: We speculate achieving lower targeted error than randomly initialized models could be sufficient. However, achieving this score is not necessary: even on exact unlearning of $S_f$, our models obtain a higher score due to samples in the retained set ($S \setminus S_f$) having noisy annotations, an unavoidable phenomena in real-world datasets. We approximate this inherent noise in the dataset by using the model retrained from scratch.

For clarity, we further describe the computation of some metrics. Our MIA has already been described in Appendix Section~\ref{sec:MIA}. Note that for measuring memorization, the deletion set is used, while for measuring generalization (a subset of) the test set is used.

\textbf{IC Targeted Error}: For the IC test between class $A$ and $B$, the targeted error represents the number of samples of class $A$ mislabelled as class $B$ and vice-versa. Intuitively, as the mislabelled samples are forgotten by the unlearning procedure, the model should confuse lesser samples between these two classes. 

\textbf{Class Removal Targeted Error}: For the class removal test removing samples from class $A$, the Targeted Error represents the number of samples the model classifies as class $A$. Intuitively, as more samples from $A$ are removed, the model should classify lesser samples into $A$. Note that if the entire class is not removed, a model that generalizes better from the partial samples still available may get penalized unnecessarily. 

\textbf{IC Error}: Error on train/test samples from the confused classes of the IC test, $A$ and $B$. 

\textbf{Class Removal Error}: Error on train/test samples of the removed class $A$.

\textbf{I.I.D confusion, Error}: Error on all samples from the train/test set. Here, a specific set of classes cannot be used for a targeted measurement.

\begin{table}[t]
\centering
\hspace{-0.25cm} 
\resizebox{0.75\textwidth}{!}{ 
 \begin{tabular}{ll|cccc} \toprule
 Method & & I.I.D Removal ($\downarrow$) & Class Removal ($\downarrow$) & I.I.D confusion ($\downarrow$) & IC ($\downarrow$) \\ \hline
 \multicolumn{6}{c}{CIFAR-10 \hspace{0.1cm} ($|S_f|=4000$)} \\\hline
 Original & & 8.4 $\pm$ 0.2 & 8.8 $\pm$ 0.4 & 14.3 $\pm$ 0.7 & 6.9 $\pm$ 0.5 \\ \hline
 \multirow{2}{*}{1-layer} & CF & 8.4 $\pm$ 0.1 & 8.3 $\pm$ 0.2 & 13.0 $\pm$ 0.9 & 6.5 $\pm$ 0.5 \\ 
 & EU & 8.5 $\pm$ 0.1 & 8.4 $\pm$ 0.3 & 12.8 $\pm$ 0.9 & 6.6 $\pm$ 0.3 \\ \cline{1-2}
 \multirow{2}{*}{10-layers} & CF & 8.4 $\pm$ 0.2 & 8.4 $\pm$ 0.2 & 12.0 $\pm$ 0.6 & 6.5 $\pm$ 0.4 \\ 
 & EU & 8.7 $\pm$ 0.1 & 8.6 $\pm$ 0.2 & 12.0 $\pm$ 0.7 & 6.7 $\pm$ 0.2 \\ \cline{1-2}
 \multirow{2}{*}{50-layers} & CF & 8.5 $\pm$ 0.1 & 8.1 $\pm$ 0.4 & 10.0 $\pm$ 0.4 & 6.1 $\pm$ 0.3 \\
 & EU & 9.3 $\pm$ 0.3 & 8.8 $\pm$ 0.4 & 10.4 $\pm$ 0.4 & 6.9 $\pm$ 0.5 \\ \hline
 Retrain & & 9.3 $\pm$ 0.1 & 8.2 $\pm$ 0.3 & 8.8 $\pm$ 0.3 & 6.4 $\pm$ 0.2 \\ \hline
  \multicolumn{6}{c}{CIFAR-100 \hspace{0.1cm} ($|S_f|=400$)} \\ \hline
 Original & & 32.1 $\pm$ 1.1 & 31.6 $\pm$ 1.1 & 32.4 $\pm$ 1.4 & 31.8 $\pm$ 0.8 \\ \hline
 \multirow{2}{*}{1-layer} & CF & 32.1 $\pm$ 1.0 & 31.7 $\pm$ 1.2 & 32.4 $\pm$ 1.3 & 31.7 $\pm$ 0.7 \\ 
 & EU & 32.1 $\pm$ 1.0 & 31.7 $\pm$ 1.1 & 32.4 $\pm$ 1.2 & 31.8 $\pm$ 0.7 \\ \cline{1-2}
 \multirow{2}{*}{10-layers} & CF & 32.4 $\pm$ 0.9 & 32.2 $\pm$ 1.1 & 32.6 $\pm$ 1.3 & 32.0 $\pm$ 0.6 \\ 
 & EU & 33.3 $\pm$ 1.2 & 32.5 $\pm$ 1.1 & 33.3 $\pm$ 1.2 & 32.8 $\pm$ 0.9 \\ \cline{1-2}
 \multirow{2}{*}{50-layers} & CF & 31.7 $\pm$ 0.9 & 31.5 $\pm$ 1.0 & 31.8 $\pm$ 0.8 & 31.2 $\pm$ 0.6 \\
 & EU & 32.1 $\pm$ 0.3 & 31.6 $\pm$ 0.2 & 31.8 $\pm$ 1.0 & 31.7 $\pm$ 1.0 \\ \hline
 Retrain & & 32.2 $\pm$ 0.4 & 31.6 $\pm$ 0.6 & 31.7 $\pm$ 1.3 & 31.8 $\pm$ 1.0 \\ 
 \bottomrule
\end{tabular}
}
\vspace{-0.25cm}
\caption{Error on the retain set distribution of test samples across unlearning tests. Scores are reported as: mean $\pm$ stdev. The EU-$k$ and CF-$k$ unlearning procedures lead to a minimal change in utility compared to retraining from scratch, unless utility is correlated with unlearning in the applied test.}
\vspace{-0.3cm}
\label{tab:utilities_across_tests}
\end{table}

\subsection{Choices for Tests}
\label{sec:TestChoices}
In the IC test we confuse samples between classes 3 (Cat) and 5 (Dog) on CIFAR10 and classes 47 (Maple Tree) and 52 (Oak Tree) on CIFAR100 unless otherwise specified. Confusing two classes can harm the overall accuracy of the original model, and we expect this effect to be more prominent when the total number of classes in the dataset is lower. The deletion set size is the same as the number of samples from one class in the training set unless otherwise specified. Note that while the size of $S_f$ is the same when comparing different tests, the size of $S_t$ is dependent on the test itself. In targeted tests (Class removal, IC), $S_t$ only has test set samples from the affected classes, whereas in untargeted tests (I.I.D Removal, I.I.D confusion) $S_t$ consists of the entire test set. In Class Removal test we remove class 0 for both CIFAR10 and CIFAR100, whereas in I.I.D Removal and I.I.D confusion we draw an equal number of samples randomly from each class.

\begin{table}[t]
 \begin{center}
 \resizebox{0.55\textwidth}{!}{ 

 \begin{tabular}{ll|cccc} \toprule
 Method & & None & Early Stop  & Cutmix & Cutmix+Early \\\hline 

\multicolumn{6}{c}{CIFAR-10 \hspace{0.1cm} ($|S_f| = 4000$)} \\\hline
 Original & & 6.53 & 7.91 & 5.81 & 8.02 \\ \hline
 \multirow{2}{*}{10-layers} & CF & 6.11 & 7.93 & 5.45 & 7.27 \\
 & EU & 6.55 & 7.88 & 5.68 & 7.08 \\ \cline{1-2}
 \multirow{2}{*}{50-layers} & CF & 5.75 & 7.31 & 5.32 & 6.75 \\
 & EU & 6.57 & 7.87 & 6.16 & 8.20 \\  \hline
 Retrain & & 6.31 & 8.50 & 5.70 & 8.97 \\ \hline

\multicolumn{6}{c}{CIFAR-100 \hspace{0.1cm} ($|S_f|=400$)} \\\hline
 Original &  & 32.53 & 33.10 & 27.26 & 30.03 \\ \hline
 \multirow{2}{*}{10-layers} & CF & 32.22 & 33.04 & 27.98 & 30.61 \\
  & EU & 33.25 & 33.23 & 28.59 & 31.23  \\ \cline{1-2}
 \multirow{2}{*}{50-layers} & CF & 30.93 & 32.37 & 27.92 & 29.37 \\
  & EU & 31.98 & 33.66 & 30.41 & 30.93 \\ \hline
 Retrain & & 30.64 & 32.62 & 26.67 & 30.67 \\ \bottomrule

\end{tabular}
}
\end{center}
\vspace{-0.3cm}\caption{Error on the retain set distribution of test samples on varying the training procedure of the original model. Regularized models have better utility even after unlearning.}\vspace{-0.4cm}
\label{tab:regularization_utilities}
\end{table}

\subsection{Utilities}

To measure utility, we compute error on unseen samples from the same distribution (unaffected classes) as $S \setminus S_f$, called the retain distribution. For the I.I.D Removal and I.I.D confusion tests, as the removal is untargeted, the evaluated samples are the same as the full test set. For the Class Removal and I.I.D confusion tests the evaluated samples consist of test set samples from the unaffected classes. This is done as error on samples from the deletion set distribution correlates with the unlearning efficacy, and thus removing them leads to a measurement of utility largely independent of unlearning.

In Table~\ref{tab:utilities_across_tests} we show the utilities of the EU-$k$ and CF-$k$ unlearning procedures across all four tests. We observe a negligible impact on utility compared to retraining from scratch, unlike most unlearning procedures suggested in existing literature. The only significant difference in error is observed in the I.I.D confusion test, where better unlearning leads to improved utility as the model gets less confused by the mislabelled samples. Note that this is not observed in the IC test as the error is reported on only the unaffected classes, where error is independent of unlearning. Thus, EU-$k$ and CF-$k$ can be used to control the unlearning-efficiency tradeoff at a fixed utility. 

In Table~\ref{tab:regularization_utilities} we show the impact of regularization on utility. We observe that early stopping slightly increases the errors, while cutmix alone reduces them especially in CIFAR100. Given the significant improvement in utility and greater downstream amenability to unlearning, using regularizers like Cutmix seems highly rewarding. Our unlearning procedures do not decrease the utility barring a slight deterioration when the training procedure uses cutmix while the unlearning procedure does not.

\begin{table}[t]
\begin{center} 
\resizebox{0.6\columnwidth}{!}{
 \begin{tabular}{ll|ccc} \toprule
 Method & Epochs & Mem & Prop. Gen. & Test-Error \\ 
  & & (Targeted Error) & (Targeted Error) & \\ \hline
  \multicolumn{5}{c}{CIFAR10} \\ \hline
 Original & - & 3016.0 & 927.0 & 16.00 \\ \hline
 \multirow{3}{*}{CF-10} & 6 & 1453 & 408 & 11.07 \\
  & 14 & 1305 & 366 & 10.63 \\ 
  & 30 & 1226 & 335 & 10.24 \\ \hline
 \multirow{3}{*}{CF-50} & 6 & 758 & 251 & 9.42 \\
  & 14 & 643 & 241 & 9.17 \\ 
  & 30 & 569 & 229 & 9.25 \\ \hline
 Retrain & 62 & 390 & 184 & 9.33 \\ \hline
 \multicolumn{5}{c}{CIFAR100} \\ \hline

 Original & - & 395 & 70 & 32.53 \\ \hline
 \multirow{4}{*}{CF-10} & 6 & 357 & 56 & 32.71 \\
  & 14 & 348 & 55 & 32.60 \\ 
  & 30 & 337 & 54 & 32.98 \\
  & 62 & 325 & 57 & 32.60 \\ \hline
 \multirow{4}{*}{CF-50} & 6 & 128 & 47 & 32.88 \\
  & 14 & 141 & 45 & 32.12 \\ 
  & 30 & 108 & 47 & 32.11 \\
  & 62 & 86 & 36 & 31.92 \\ \hline
 Retrain & 126 & 64 & 31 & 30.82 \\
 \bottomrule
\end{tabular}}
\end{center}
\vspace{-0.35cm}
\caption{Varying catastrophic forgetting epochs on the IC test. The number of epochs used for fine-tuning can further control the forgetting-efficiency tradeoff without hurting utility.}
\label{tab:varyingCFepochs}
\vspace{-0.5cm}
\end{table}

\begin{figure*}[t]
\hspace{-1.1cm}
\includegraphics[width=1.1\columnwidth]{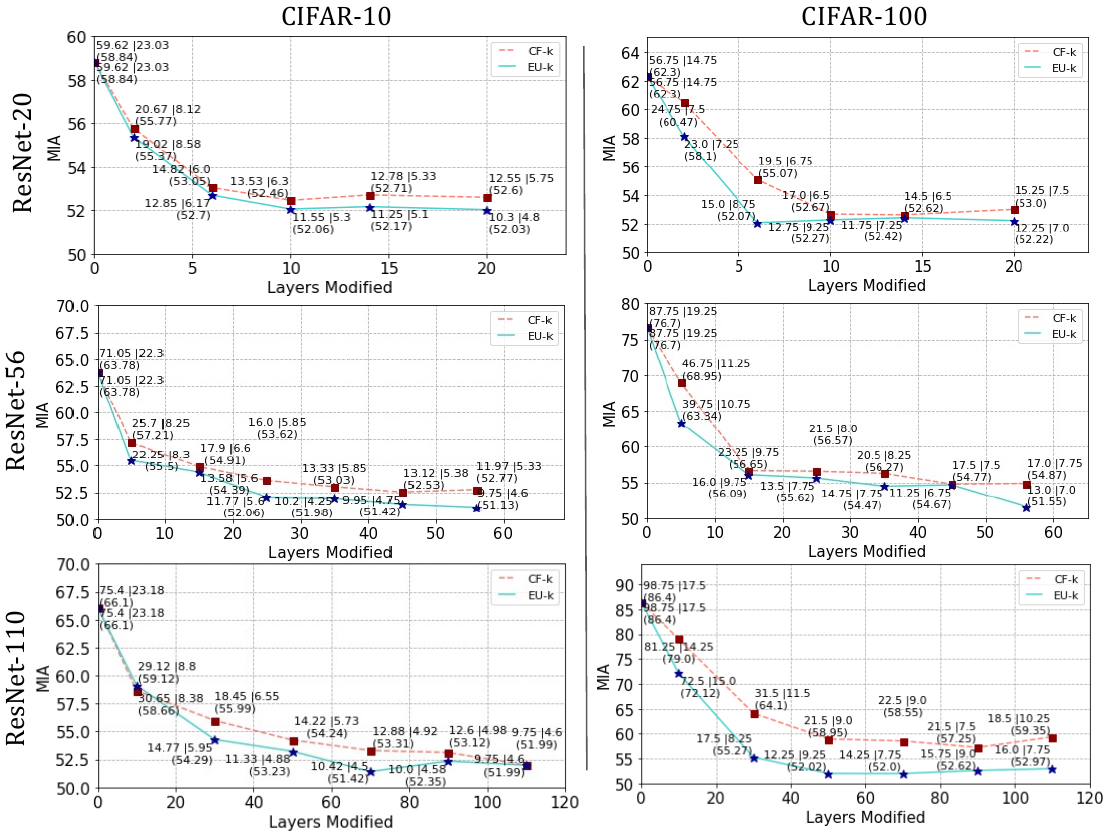}
\caption{We plot the MIA (Y) vs number of layers unlearnt using EU-$k$ (solid blue) and CF-$k$ (dashed red) for different architectures across datasets. For each model (point) we report three forgetting metrics as `memorization $|$ property generalization (MIA)' with memorization and property generalization computed using targeted error. The leftmost point is the original model while the rightmost EU point is the full retrained model. We observe consistent observations with the main paper across metrics and datasets.}
\label{fig:varlayers}
\end{figure*}

\section{Additional Experiments}

Finally, we vary some of the choices we make in our experiments to demonstrate the robustness of our observations.

\subsection{Varying the number of unlearning epochs}
The original experiments train CF-$k$ models for half the epochs compared to EU-$k$ models. In Table~\ref{tab:varyingCFepochs} we compare the variation of performance among CF-$k$ models at the end of each warm restart while finetuning. While less information is unlearnt on reducing epochs, even six epochs are sufficient for drastic improvements in forgetting, with no significant change in utility (error on full test set). The number of catastrophic forgetting epochs can thus be reduced, and control the forgetting-efficiency tradeoff at constant utility.

\subsection{Varying the number of layers}
\label{sec:varlayers}
In Figure~\ref{fig:varlayers} we show results of varying $k$ for 3 different ResNet depths: 20, 56 and 110. The IC test is able to detect retained information despite exact unlearning of almost 30\% of the final layers. However, on unlearning the final half of the network, its unclear whether most information is removed or the IC test is unable to identify the presence of retained information. CF-$k$ is consistently within a small margin of EU-$k$ demonstrating the catastrophic forgetting is able to lose enough information to match EU while being two times faster.

\subsection{Varying Amount of Untargeted Removal}
\begin{figure*}[t]
\centering
\includegraphics[width=\columnwidth]{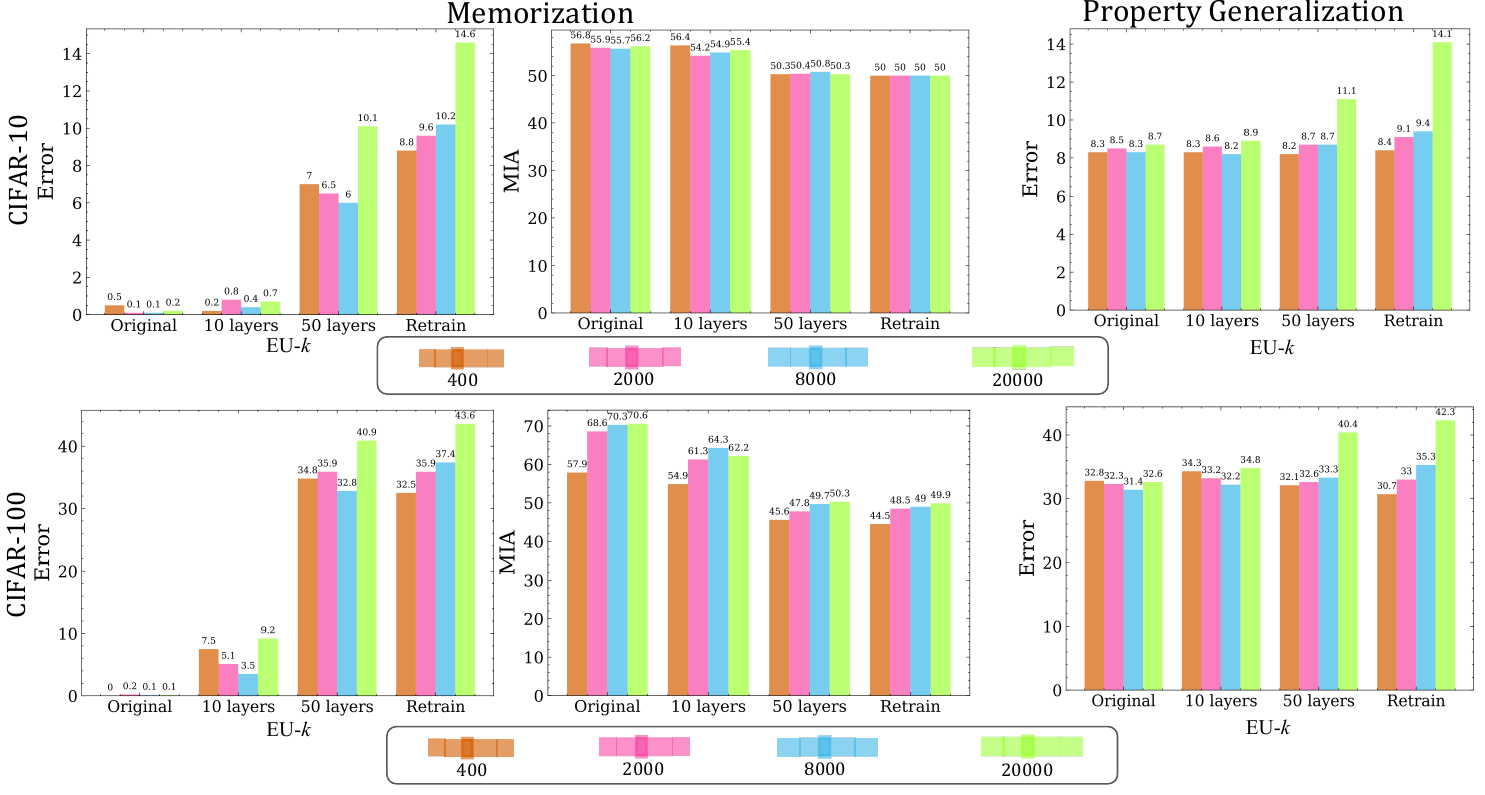}
\caption{Varying $|S_f|$ for I.I.D Removal test. Error seems to distinguish varying levels of memorization, but needs huge deletion sets (50\% of dataset size) in the case of property generalization. Moreover, here error has the limitation of misaligning forgetting ($\uparrow$ is better) and utility ($\downarrow$ is better).}
\label{fig:RStest}
\end{figure*}
\begin{figure*}[t]
\centering
\includegraphics[width=\columnwidth]{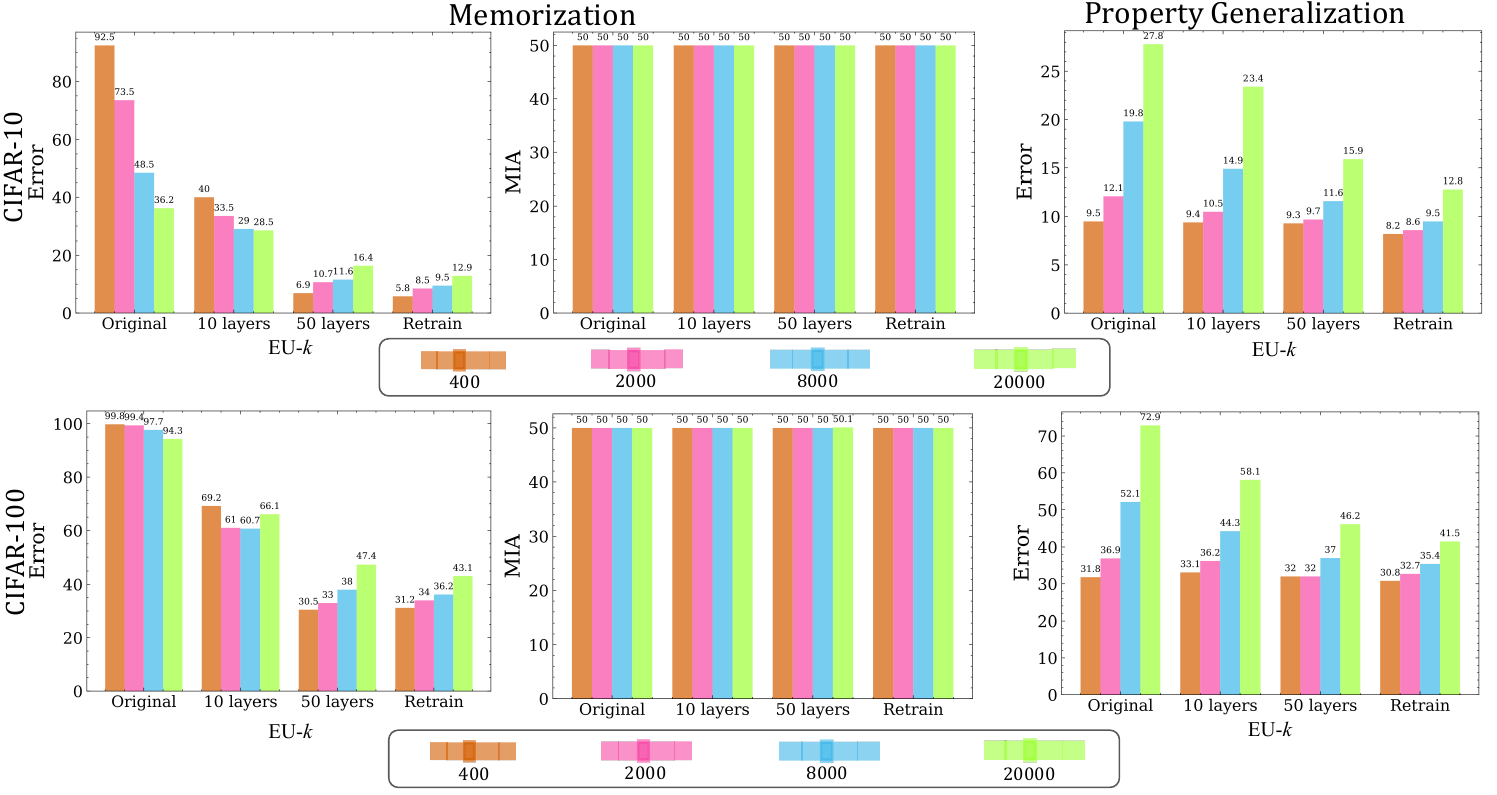}
\caption{Varying $|S_f|$ for I.I.D Confusion test. Error reliably measures memorization even in small deletion sets (1\% of deletion set size), though much larger ones (20\% of deletion set size) are needed to produce detectable effects on property generalization.}
\label{fig:RCtest}
\end{figure*}
\label{sec:untargetedvarying}
In Figures~\ref{fig:RStest} and \ref{fig:RCtest}, we show the forgetting performance when we vary deletion set sizes in tests with untargeted removal: I.I.D Removal and I.I.D Confusion. Here, we use larger sizes than those reported in the main paper as smaller deletion sets show negligible trends in untargeted removal. For detecting effects on property generalization, Error on I.I.D confusion test needs far fewer samples than Error on I.I.D Removal. For memorization, we see that Error is able to distinguish and rank models fairly well whereas MIA works well in the case of I.I.D Removal test but fails completely on the I.I.D Confusion test. CF models continue to be close to EU models here and the gap between them decreases as we add more confusion. Overall, untargeted removal requires much larger deletion sets to show clear forgetting trends as compared to targeted removal, demonstrating the usefulness of strategic sampling.

\subsection{Varying Amount of Targeted Removal}
\begin{figure*}[t]
\centering
\includegraphics[width=\columnwidth]{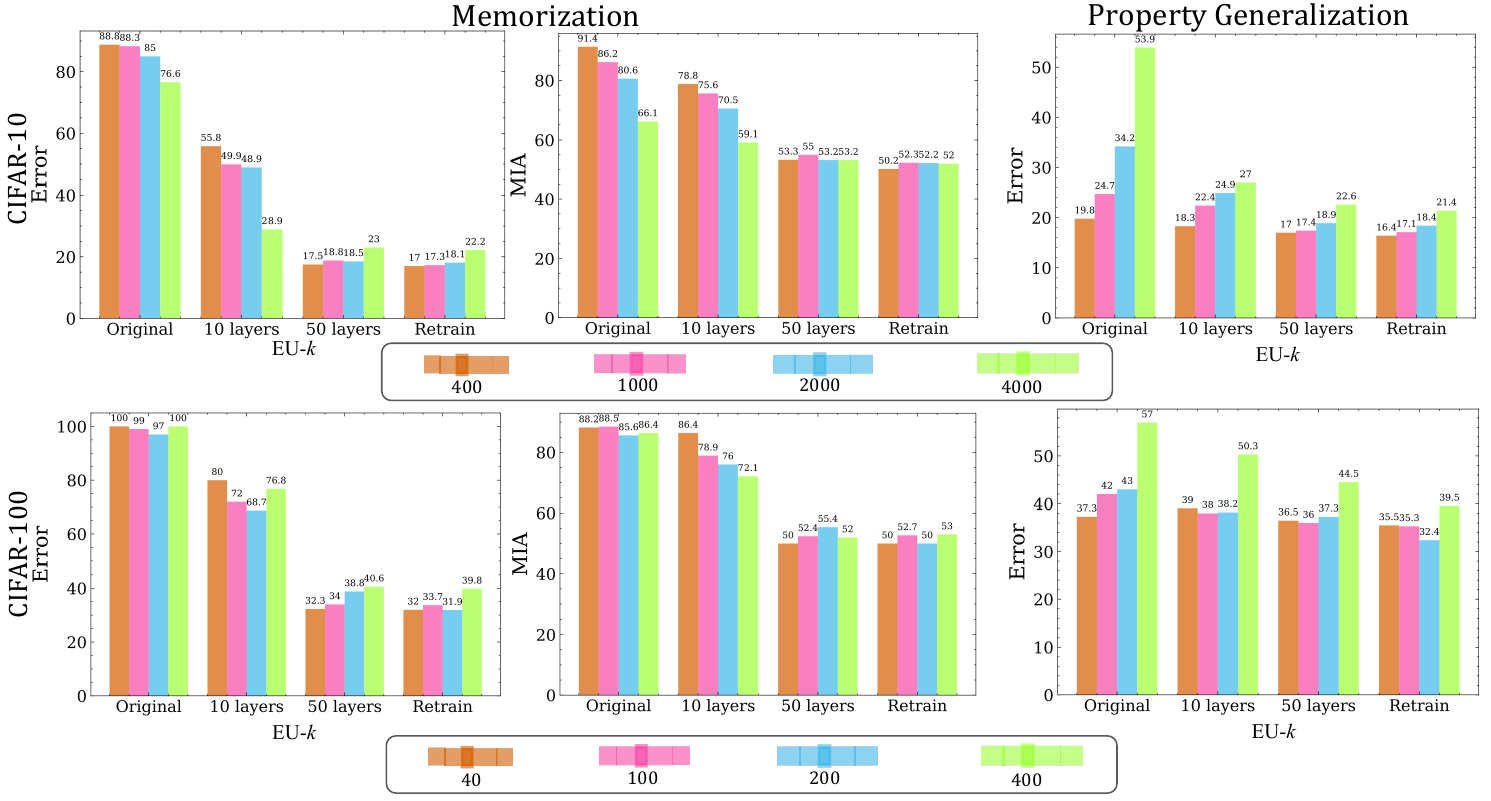}
\caption{Varying $|S_f|$ for IC test. In CIFAR10, at 1\% of dataset size, the IC test reliably detects imperfect forgetting across metrics. In CIFAR100, imperfect removal of memorization is detected at 1\% of the class size, a noticeable effect on generalization requires a larger deletion set (5\% of dataset size).}
\label{fig:ICtest}
\end{figure*}
\begin{figure*}[t]
\centering
\includegraphics[width=\columnwidth]{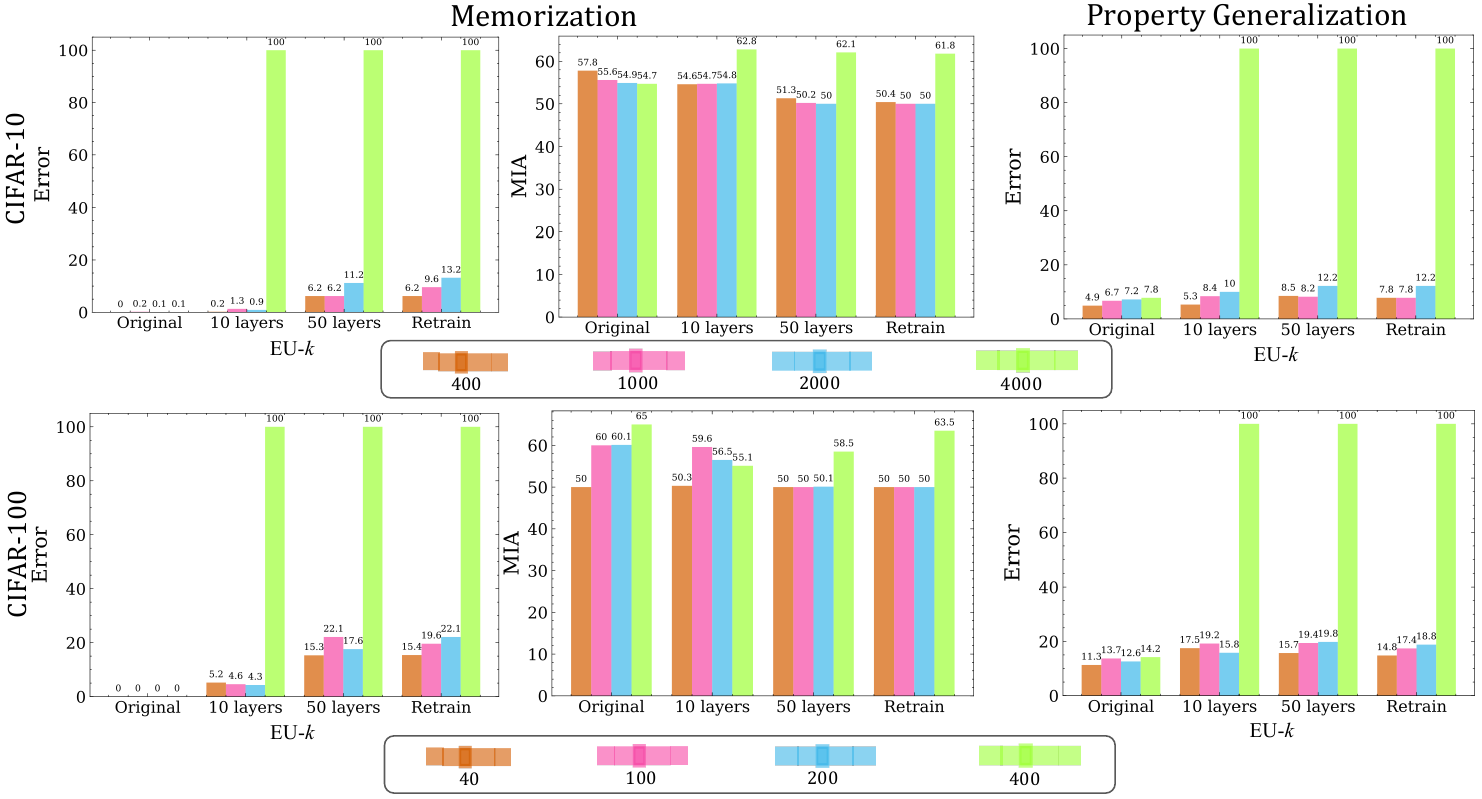}
\caption{Varying $|S_f|$ for Class removal test. The Class removal test is not able to reliably distinguish varying levels of property generalization and provides a weak signal for memorization, particularly for small $|S_f|$.}
\label{fig:CRtest}
\end{figure*}
\label{sec:targetedvarying}
Now, we study the forgetting performance for partial Class Removal and partial Interclass Confusion. We show results for varying $|S_f|$ from 10\% samples of a class to the size of an entire class (as used in the original paper).

First, we present the results of the IC test in Figure~\ref{fig:ICtest}. We see that for memorization all metrics are reflective even when a very small subset of samples is confused. Error and MIA having increasingly better contrast for smaller deletion sets.

Then, we present the results of the Class Removal test in Figure~ \ref{fig:CRtest}. The Class Removal test has significantly different behavior when all samples of the class are removed compared to partial class removal. In the case of full Class Removal, all information about the class is removed, and hence an unlearnt model is expected to not classify any sample as the removed class. However, in partial Class Removal, a well generalized model may correctly classify more samples as the affected class, thus leading to the misalignment of utility and forgetting. We observe that MIA seems to have unclear trends in partial Class Removal, sometimes giving a weak signal for unlearning efficacy. 

\subsection{Varying Confused Classes}
\label{sec:varclasses}
\begin{figure*}[t]
\centering
\includegraphics[width=0.85\columnwidth]{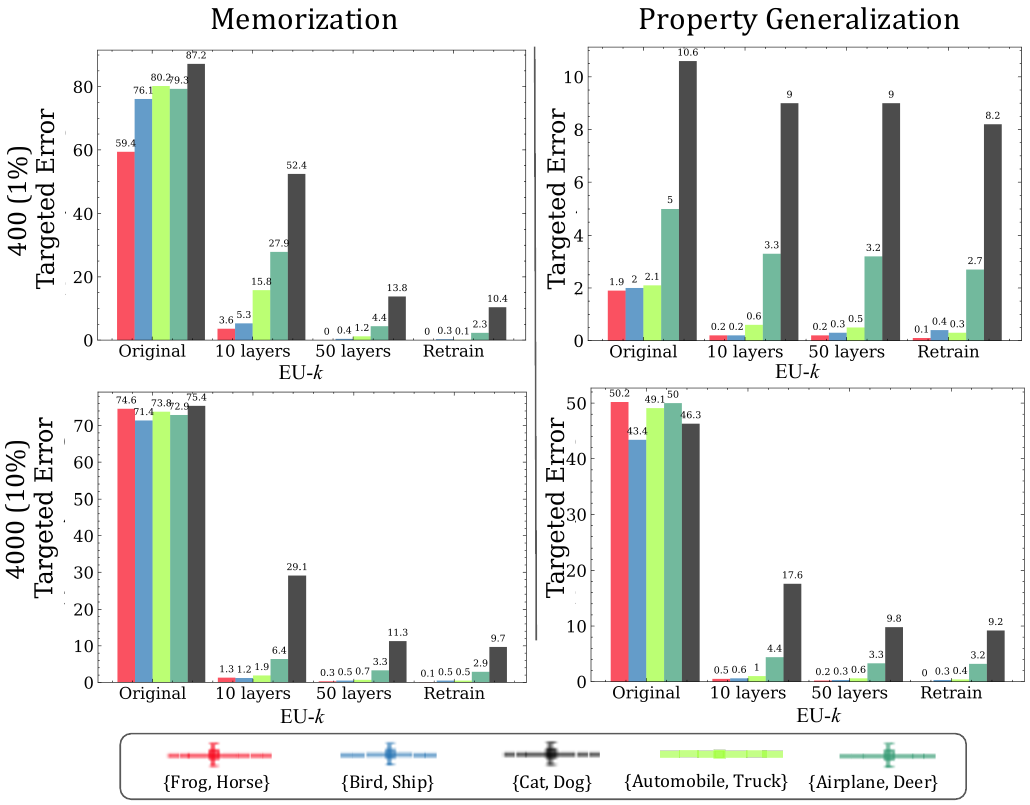}
\caption{Varying confused class pairs on CIFAR10, with the similarity of the classes increasing from left to right in each group of bars. While the IC test reliably detects imperfect forgetting across class pairs, the trends are clearer for more similar classes.}
\label{fig:varclasses}
\end{figure*}

Throughout our experiments, we only confused the hardest pair of classes in the dataset (Cat and Dog for CIFAR10, Maple Tree and Oak Tree for CIFAR100). In Figure~\ref{fig:varclasses} we ablate the chosen class pair, grouping the ten classes in CIFAR10 into five pairs to maximize diversity. The five pairs are arranged in increasing order of similarity below along with their bar color:

\begin{itemize}
    \item Frog (6) - Horse (7): Red
    \item Bird (2) - Ship (8): Blue
    \item Airplane (0) - Deer (4): Light Green
    \item Automobile (1) - Truck (9): Dark Green
    \item Cat (3) - Dog (5): Black
\end{itemize}

We can see that the number of confused samples by any model is much higher as we go from left to right, indicating that confusing a similar pair of classes makes unlearning more difficult. Both memorization and property generalization trends across varying levels of unlearning, from Original to Retrain, are consistently preserved. This shows that irrespective of the chosen class pair, the IC test is able to clearly distinguish varying degrees of forgetting. 

\subsection{Learning Without Restarts}

\begin{table}
\centering

\resizebox{0.5\columnwidth}{!}{ 
 \begin{tabular}{l|cc|cc} \toprule
Sched & \multicolumn{2}{c|}{CF-$10$} & \multicolumn{2}{c}{CF-$50$}\\ \cline{2-5}
& MIA & Targeted Error & MIA & Targeted Error \\ \hline
 \multicolumn{5}{c}{CIFAR10 \hspace{0.1cm} ($|S_f|=4000$)} \\ \hline
WR & 58.66 & 335 & 54.24 & 229 \\
No & 58.10 & 340 & 53.62 & 234 \\ \hline
 \multicolumn{5}{c}{CIFAR100 \hspace{0.1cm} ($|S_f|=400$)} \\ \hline
WR & 77.99 & 58 & 58.24 & 41 \\
No & 78.87 & 55 & 58.57 & 43 \\
 \bottomrule
\end{tabular}}
\vspace*{-0.25cm}
\caption{We compare Warm Restarts and keeping a single learning rate cycle between the same maxLR and minLR. MIA represents memorization while Targeted Error measures property generalization.}
\label{tab:norestarts}
\vspace*{-0.25cm}
\end{table}

One concern which may arise is whether catastrophic forgetting performs well due to warm restarts in our learning rate schedule. We ablate this effect in Table \ref{tab:norestarts} and see that in all cases removing warm restarts has no effect on the degree of catastrophic forgetting. 

\end{document}